%% file: main.tex
\documentclass[sigconf]{aamas} 

\renewcommand\footnotetextcopyrightpermission[1]{} 
\usepackage{balance} 

\usepackage{multirow} 

\usepackage{subcaption}

\newcommand{\xhdr}[1]{\vspace{5pt}\noindent\textbf{#1 }}
\newcommand{\ignore}[1]{}

\newcommand{\argmax}{\operatornamewithlimits{argmax}}

\newcommand{\squishlist}{\begin{list}{$\bullet$}{\topsep=1pt \parsep=0pt \itemsep=1pt \leftmargin=1em }} 
\newcommand{\squishend}{\end{list}}

\newcommand{\beitemize}{\begin{list}{$\bullet$}{}} 
\newcommand{\enitemize}{\end{list}}

\theoremstyle{plain}
\newtheorem{theorem}{Theorem}

\newtheorem{corollary}{Corollary}

\theoremstyle{definition}
\newtheorem{definition}{Definition}
\theoremstyle{remark}

\newcommand{\Z}{\mathbf{Z}} 

\setcopyright{ifaamas}
\acmConference[AAMAS '26]{Proc.\@ of the 25th International Conference
on Autonomous Agents and Multiagent Systems (AAMAS 2026)}{May 25 -- 29, 2026}
{Paphos, Cyprus}{C.~Amato, L.~Dennis, V.~Mascardi, J.~Thangarajah (eds.)}
\copyrightyear{2026}
\acmYear{2026}
\acmDOI{}
\acmPrice{}
\acmISBN{}

\acmSubmissionID{1511}

\title[Past Discounting]{Past-Discounting is Key for Learning Markovian Fairness with Long Horizons}

\author{Ashwin Kumar}
\affiliation{
  \institution{Washington University in St Louis}
  \city{St Louis, MO}
  \country{USA}}
\email{ashwinkumar@wustl.edu}

\author{William Yeoh}
\affiliation{
  \institution{Washington University in St Louis}
  \city{St Louis, MO}
  \country{USA}}
\email{wyeoh@wustl.edu}

\begin{abstract}

Fairness is an important consideration for dynamic resource allocation in multi-agent systems. 
Many existing methods treat fairness as a one-shot problem without considering temporal dynamics, which
misses the nuances of accumulating inequalities over time. Recent approaches overcome this limitation by tracking allocations over time, assuming perfect recall of all past utilities. While the former neglects long-term equity, the latter introduces a critical challenge: the augmented state space required to track cumulative utilities grows unboundedly with time, hindering the scalability and convergence of learning algorithms. Motivated by behavioral insights that human fairness judgments discount distant events, we introduce a framework for temporal fairness that incorporates past-discounting into the learning problem.  This approach offers a principled interpolation between instantaneous and perfect-recall fairness. Our central contribution is a past-discounted framework for memory tracking and a theoretical analysis of fairness memories, showing past-discounting guarantees a bounded, horizon-independent state space, a property that we prove perfect-recall methods lack. This result unlocks the ability to learn fair policies tractably over arbitrarily long horizons. We formalize this framework, demonstrate its necessity with experiments showing that perfect recall fails where past-discounting succeeds, and provide a clear path toward building scalable and equitable resource allocation systems. 

\end{abstract}

\keywords{Fairness, Resource Allocation, Long-horizon Learning }

\newcommand{\BibTeX}{\rm B\kern-.05em{\sc i\kern-.025em b}\kern-.08em\TeX}

\begin{document}


\pagestyle{fancy}
\fancyhead{}


\maketitle 

\section{Introduction}

Dynamic resource allocation is central to many real-world applications, from matching passengers to taxis~\cite{shah2020neural,ride_alonso} and distributing aid to the homeless~\cite{kube2019allocating}, to allocating life-saving vaccines~\cite{mcilrathNonMarkovianFairness}. In these long-horizon settings, agents must balance efficiency with fairness, but existing approaches are caught in a fundamental dilemma. \textbf{Myopic fairness} approaches look at fairness within a single allocation step, ignoring any past allocations and leading to cumulative inequity. To overcome this limitation, recent works tackle temporal fairness by tracking cumulative utilities over time~\cite{jiang2019FEN,zimmer2021MOMDP,mcilrathNonMarkovianFairness}—what we term \textbf{perfect-recall fairness}—but these introduce a critical and previously unaddressed failure mode. As noted by~\citet{mcilrathNonMarkovianFairness}, maintaining fairness with history dependence requires augmenting the state. However, we show that with perfect recall, this augmented state space grows unboundedly with time, rendering learning algorithms computationally intractable and preventing them from converging on fair policies.

How can an agent remember the past without being crippled by it? We find inspiration in behavioral economics and moral psychology, which show that human fairness judgments naturally discount distant events. People perceive events in the distant past more abstractly~\cite{trope2010construal}, devalue temporally removed outcomes~\cite{frederick2002time}, and are more inclined to forgive past transgressions as time passes~\cite{li2021close,wohl2007perception}. Motivated by these insights, we introduce a framework that incorporates \textbf{past-discounting}. By applying a tunable discount factor to historical utilities, our approach provides a principled interpolation between flawed instantaneous fairness and intractable perfect-recall fairness. This is not merely a heuristic; we prove that past-discounting is the key to achieving what was previously impossible: \textbf{Markovian fairness with a provably bounded and horizon-independent state space.} This theoretical guarantee ensures the computational tractability required for reinforcement learning agents to learn robustly fair policies, no matter how long the task horizon.

\xhdr{Contributions:} Our contributions are:
\begin{enumerate}
    \item We identify and formalize a critical failure mode in long-term fairness methods~\cite{mcilrathNonMarkovianFairness,zimmer2021MOMDP}: \textbf{perfect-recall causes unbounded state-space growth}, rendering learning intractable over long horizons.
    \item We propose \textbf{past-discounted fairness memories}, a novel, behaviorally-grounded framework that balances historical context with computational feasibility by applying a geometrically decaying weight to past utilities.
    \item We provide a \textbf{rigorous theoretical proof} that our past-discounting mechanism guarantees a bounded and horizon-independent augmented state space, retaining the Markov property while enabling tractable learning.
    \item We empirically \textbf{demonstrate the necessity of discounting the past}, showing that an RL agent successfully learns fair policies with our method while systematically failing with perfect recall as the time horizon increases.
\end{enumerate}
\section{Related Work}

Fairness in resource allocation has a rich history, but traditional literature has focused on static, one-shot problems. Concepts like proportionality, envy-freeness, and maximin share guarantees~\cite{budish2011combinatorial,procaccia2014fair} provide robust solutions for individual decision points. However, their myopic nature makes them insufficient for dynamic settings like taxi matching or aid distribution, where  long-term equity necessitates reasoning over multiple time steps. Even in the sequential setting, many approaches consider fairness over only the current valuations of resources or considering future value, but do not consider past memories~\cite{sinclair2022sequential, hosseini2024class}. We call this \textbf{myopic} or \textbf{instantaneous fairness}.

Recognizing this, recent research has explored temporal fairness across various domains. In multi-agent reinforcement learning (MARL), some approaches constrain per-step allocations or evaluate cumulative utilities at a terminal stage~\cite{jiang2019FEN,zimmer2021MOMDP}. More advanced mechanisms include forecasting for ride-hailing applications~\cite{kumar2023SI} and hierarchical frameworks to mediate between efficiency and fairness~\cite{jiang2019FEN}. Similar challenges are addressed in sequential voting~\cite{lackner2020perpetual} and online fair division~\cite{aleksandrov2015online}, where a series of interdependent choices necessitates a long-term perspective. These works establish a clear consensus: achieving meaningful fairness in sequential settings requires memory of the past.

The dominant paradigm for incorporating this memory is \textbf{perfect recall}, where agents track the complete sum of historical utilities~\cite{zimmer2021MOMDP, siddique2020MOMDP, mcilrathNonMarkovianFairness, jiang2019FEN, cookson2025temporal, lodi2024fairness}. The recent work by \citet{mcilrathNonMarkovianFairness} formalizes this by treating history-dependent fairness as a non-Markovian problem. They show that by augmenting the state with a memory of past utilities (i.e., perfect recall), one can restore the Markov property. However, this approach introduces a critical, unaddressed challenge: for a perfect memory, the augmented state space grows unboundedly with the time horizon. This makes learning computationally intractable, especially in long-horizon tasks. Our work directly addresses this limitation: \textbf{we seek a memory mechanism that is not only Markovian but also improves tractability.}

This leaves the field caught between two extremes: myopic fairness, which is computationally simple but ignores long-term equity (especially over past allocations), and perfect-recall fairness, which is equitable in theory but can be computationally infeasible, as we show later. Our work introduces a third paradigm that occupies the practical middle ground: \textbf{past-discounted fairness}. This approach is motivated by strong evidence from behavioral economics and moral psychology, which shows that human fairness judgments are sensitive to temporal distance. People naturally devalue outcomes that are further in the past~\cite{trope2010construal,frederick2002time} and are more forgiving of older transgressions~\cite{li2021close,wohl2007perception}. This concept is analogous to future-discounting in standard reinforcement learning~\cite{mnih2013playing,sutton1998introduction}, but its application to past-oriented fairness memories to ensure tractability remains unexplored.

In summary, while prior work has correctly identified the need for memory in temporal fairness, it has not provided a computationally viable solution for long-horizon problems. Our contribution is to formalize a behaviorally-grounded, past-discounted memory that is (i) sufficient to restore the Markov property with a bounded, horizon-independent state, and (ii) provides a practical mechanism for learning fair policies in complex, long-running systems. To the best of our knowledge, no existing work has considered using past discounts to learn fair behavior.

\section{Preliminaries}

Social welfare functions provide a mathematical formulation to evaluate both fairness and efficiency in resource allocation. Given a utility vector 
\begin{equation}
\Z = (z_1, z_2, \ldots, z_n)
\end{equation}
representing the utilities received by $n$ agents, many social welfare functions have been considered in the literature, including:

\begin{itemize}
    \item \textbf{Utilitarian Welfare}, 
    which maximizes the total utility without explicit fairness considerations~\cite{sen2017collective}.
    \begin{equation}
        W_U(\Z) = \sum_{i=1}^{n} z_i,
    \end{equation}
    
    \item \textbf{Egalitarian Welfare}, 
    which prioritizes the well-being of the worst-off agent. This is also known as Rawlsian or maximin fairness~\cite{rawls1971atheory}.
    \begin{equation}
        W_{MMF}(\Z) = \min_{i} z_i,
    \end{equation}
    
    \item \textbf{Nash Welfare}, 
    which balances fairness and efficiency. This measure is rooted in Nash's bargaining solution~\cite{nash1950bargaining} and has been influential in fair division research~\cite{caragiannis2019unreasonable}.
    \begin{equation}
        W_N(\Z) = \prod_{i=1}^{n} z_i,
    \end{equation}
    
\end{itemize}

Traditionally, these functions evaluate fairness at a single point in time, thereby ignoring the history of past allocations and expectations for future ones. In dynamic settings, several approaches extend these welfare functions by incorporating historical and predictive elements. For example, some works cast the allocation problem as a Multi-Agent Reinforcement Learning (MARL) task that optimizes fairness at an intermediate or terminal state~\cite{jiang2019FEN,zimmer2021MOMDP,siddique2020MOMDP, kumar2025decaffull}, while others employ Non-Markovian Decision Processes that explicitly account for the entire past trajectory of allocations~\cite{mcilrathNonMarkovianFairness}.

In many formulations, the allocation at time $t$, denoted by $\mathcal{A}^t$, is defined as a mapping of resources to agents such that $\mathcal{A}^t_i$ represents the resources allocated to agent $i$. The welfare corresponding to the post-allocation utility vector $\Z^t|\mathcal{A}^t$ is then given by $W(\Z^t|\mathcal{A}^t)$, and the optimal allocation is defined as:
\begin{align}
    \mathcal{A}^{t*} &= \argmax_{\mathcal{A}} W(\Z^t|\mathcal{A}). \label{eq:Opt}
\end{align}
Here, we denote by $u^\mathcal{A}_i$ the utility derived by agent $i$ from allocation $\mathcal{A}$, and by $u^t_i$ the utility actually received by agent $i$ at time $t$.

There exist various methods to compute the utility vector $\Z^t|\mathcal{A}$, and these choices influence the resulting allocation. We discuss some popular approaches below, motivating and building up to past-discounted fairness.

\begin{figure*}[t]
    \centering
    \includegraphics[width=0.3\linewidth]{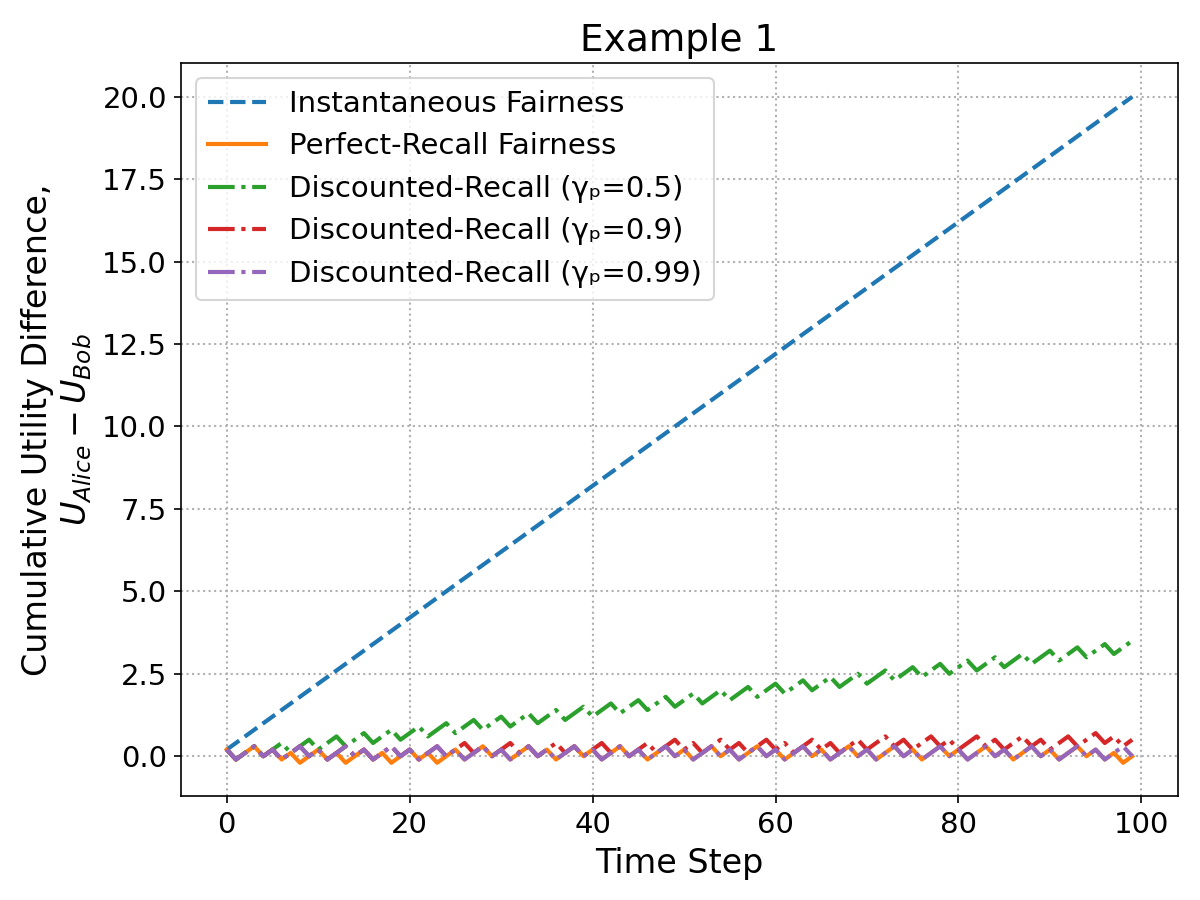}
    \includegraphics[width=0.3\linewidth]{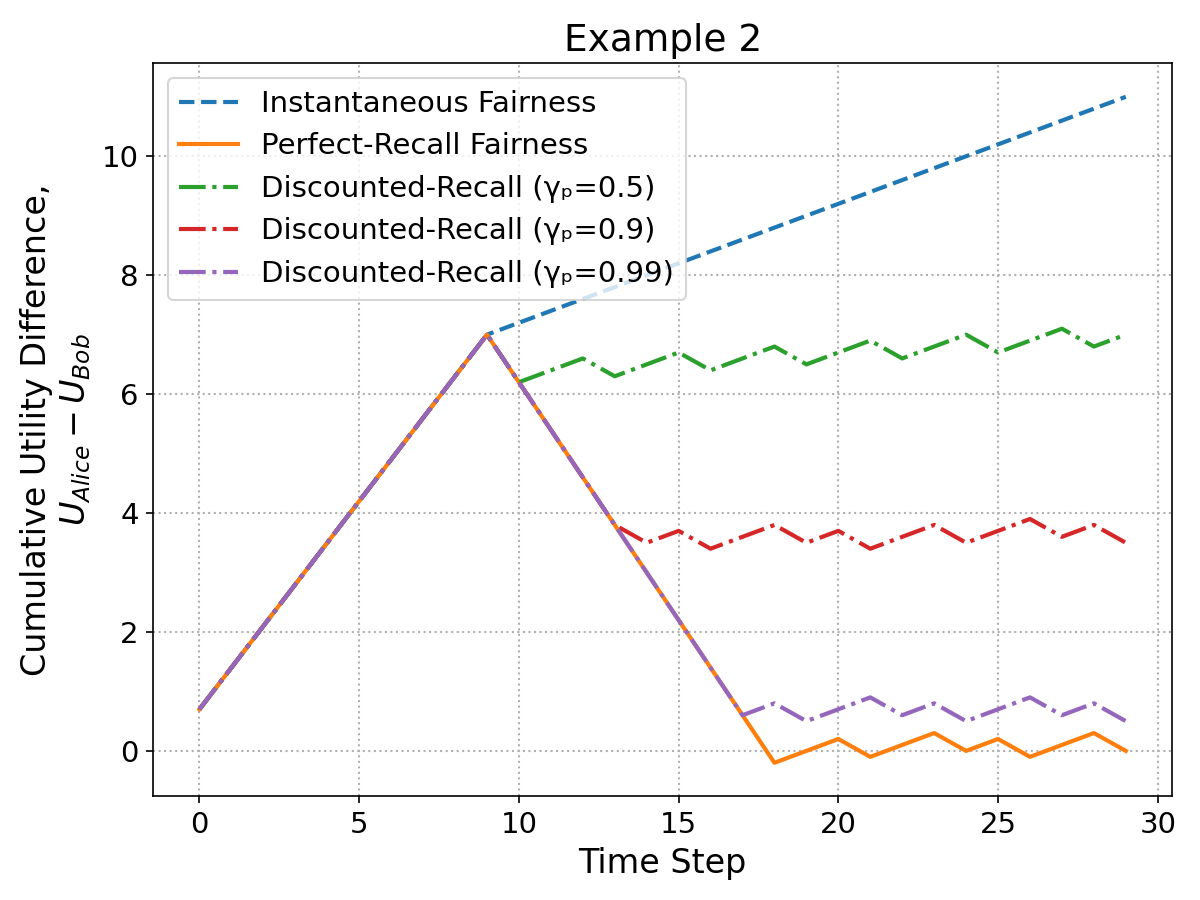}
    \includegraphics[width=0.3\linewidth]{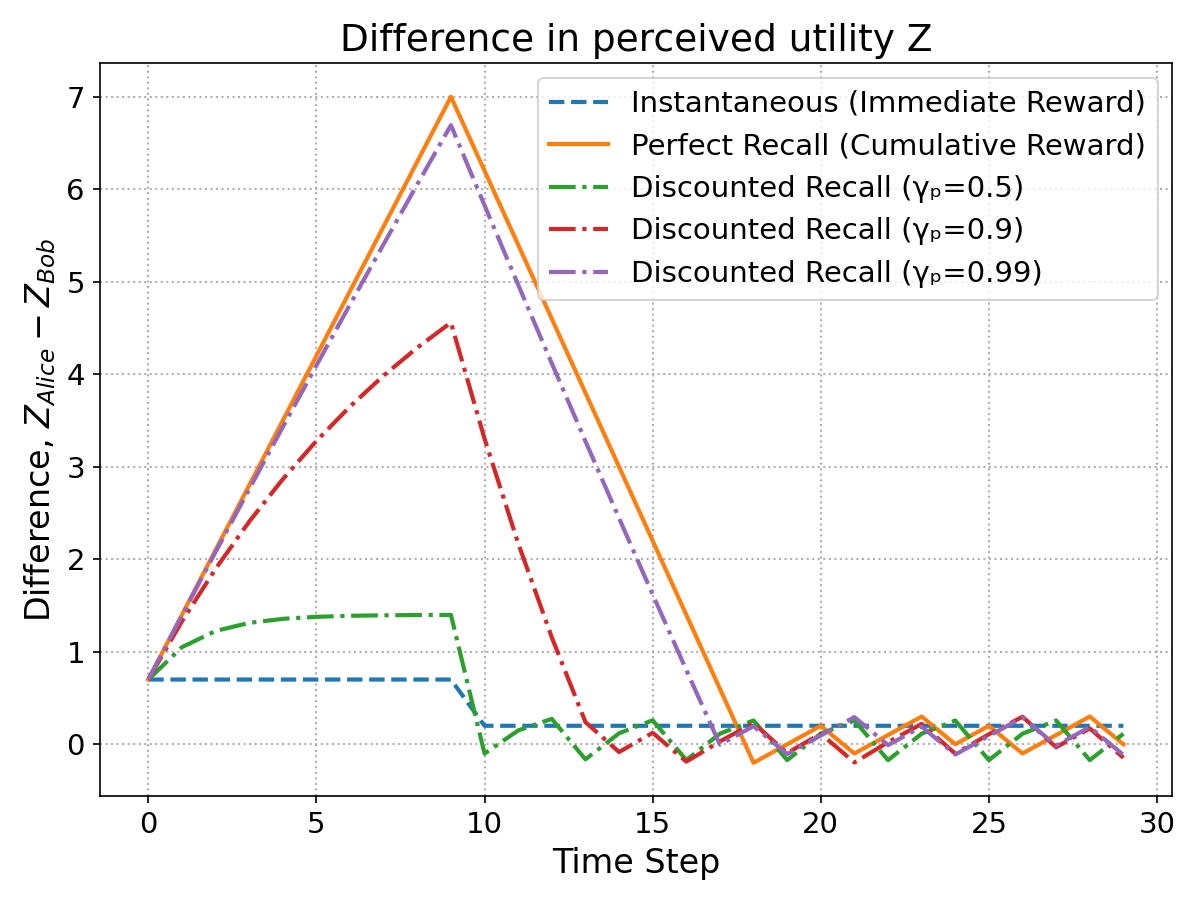}
    \caption{Comparison of cumulative utility differences under different fairness paradigms. 
    (Left) Cumulative utility difference, $\sum U_{Alice} - \sum U_{Bob}$, over time for Example~\ref{ex:inst_example}, where both agents participate from the start. 
    (Center) Cumulative utility difference, $\sum U_{Alice} - \sum U_{Bob}$, over time for Example~\ref{ex:recall_example},  where only Alice is active initially and Bob joins later. 
    (Right) Difference in perceived utility between Alice and Bob for all three methods for Example~\ref{ex:recall_example}. This plot shows the effect of $\gamma_p$ on the perceived values. It changes the speed at which we forget past decisions, interpolating between perfect-recall and instantaneous fairness.
    }
    \label{fig:combined}
\end{figure*}

\section{Temporal Fairness in Resource Allocation}

In dynamic resource allocation, fairness must be evaluated not only on the basis of the current decision but also by considering past allocations and future expectations. In this section, we present three paradigms for temporal fairness: \emph{Instantaneous fairness}, \emph{perfect-recall historical fairness}, and our proposed \emph{discounted-recall historical fairness}. We define each approach, illustrate them with examples, and discuss their inherent limitations.

\subsection{Instantaneous Fairness}

Instantaneous fairness considers only the current allocation decision. Formally:
\begin{align}
    Z^t_i \mid \mathcal{A} = u^\mathcal{A}_i,
\end{align}
which implies that fairness is assessed solely on the utility \(u^\mathcal{A}_i\) derived from the current allocation. In this formulation, the welfare function \(W\) is optimized based solely on the immediate utilities, yielding an allocation that is deemed fair at that specific time step.

\begin{definition}[Instantaneous Fairness]\label{def:instantaneous_fairness}
An allocation exhibits \emph{instantaneous fairness} at time \(t\) if fairness is evaluated solely on the one-step utilities. For any candidate allocation \(\mathcal{A}\), define
\begin{align}
    \Z^t \mid \mathcal{A} = \bigl(u^\mathcal{A}_1,\ldots,u^\mathcal{A}_n\bigr).
\end{align}
Then an instantaneous-fair allocation at time \(t\) is any optimizer
that
selects the allocation that maximizes the welfare applied to the current-step utility vector.
\end{definition}

Although this approach often produces a solution that is optimal for that particular time step, it neglects the temporal dimension by ignoring both historical allocations and anticipated future resources. For example, consider:

\begin{example}\label{ex:inst_example}
Two agents, Alice and Bob, compete for two indivisible items: a cake and a donut. Suppose 
\[
\text{Alice: }(u_{\text{cake}}, u_{\text{donut}}) = (0.2, 0.5), \quad
\text{Bob: }(0.3, 0.5).
\]
In a purely instantaneous allocation, the donut is assigned to the agent who slightly benefits from it more in that step (Alice, compared to the baseline 0.2). Over repeated interactions, however, Alice may receive a disproportionate number of donuts, leading to a cumulative imbalance.
\end{example}

Thus, while instantaneous fairness might maximize short-term objectives, its disregard for temporal dynamics can result in significant long-term disparities.

\subsection{Perfect-Recall Historical Fairness}

To capture the temporal aspect of fairness, perfect-recall historical fairness incorporates all past allocations into the fairness evaluation. Instead of relying solely on the instantaneous utility, we define an adjusted utility vector \(\Z^t\) that aggregates utilities over all previous steps:
\begin{align}
    Z^t_i = \sum_{\tau=0}^{t} u_i^\tau.
\end{align}
Consequently, the post-allocation utility becomes:
\begin{align}
    Z^t_i \mid \mathcal{A} &= \sum_{\tau=0}^{t-1} u^\tau_i + u^\mathcal{A}_i \\
    &= Z^{t-1}_i + u^\mathcal{A}_i.
\end{align}
In some cases, averaging these utilities over time may be preferable:
\begin{align}
    Z^t_i \mid \mathcal{A} = \frac{Z^{t-1}_i \cdot (t-1) + u^\mathcal{A}_i}{t}.
\end{align}

\begin{definition}[Perfect-Recall Fairness]\label{def:perfect_recall_fairness}
An allocation exhibits \emph{perfect-recall fairness} at time \(t\) if fairness is evaluated using all past allocations. For any candidate allocation \(\mathcal{A}\), define \(\Z^t \mid \mathcal{A}\) by either:
\begin{align}
    \text{(cumulative)}\quad & Z^t_i \mid \mathcal{A} = Z^{t-1}_i + u^\mathcal{A}_i, \\
    \text{(averaged)}\quad & Z^t_i \mid \mathcal{A} = \frac{Z^{t-1}_i \cdot (t-1) + u^\mathcal{A}_i}{t}.
\end{align}
A perfect-recall-fair allocation at time \(t\) is any optimizer
that
selects the allocation that maximizes the welfare applied to the history-aggregated utility vector.
\end{definition}

This approach is especially relevant in domains such as long-term healthcare or education funding, where addressing historical disparities is crucial. However, perfect recall may overcompensate past imbalances. For instance:

\begin{example}\label{ex:recall_example}
Suppose Alice is the sole participant for 10 steps and accumulates a high utility. If Bob joins at step 11, perfect-recall fairness might allocate many future resources to Bob to ``catch him up.'' This could be viewed as unfair to Alice, as her early contributions—made in Bob's absence—should not overly penalize her in future allocations.
\end{example}

\subsection{Past-Discount: Discounted-Recall Historical Fairness}

To balance the extremes of instantaneous and perfect-recall fairness, we propose \emph{discounted-recall historical fairness}. This paradigm introduces a temporal decay factor \(\gamma_p \in [0,1]\) that gradually diminishes the influence of older allocations. The intuition is that while historical context is important, its influence should naturally decay over time. Such a decay mechanism is inspired by behavioral research, which indicates that humans discount temporally distant events. Furthermore, incorporating past discounts aligns how we consider past utilities with how future rewards are treated in sequential decision-making (SDM), where temporal decay is crucial for ensuring convergence of returns and for tractable computation.

\subsubsection{Discounted Recall with Additive Utilities}

In the additive setting, past utilities are discounted and then combined with the current utility:
\begin{align}
    Z^t_i \mid \mathcal{A} = \gamma_p\, Z^{t-1}_i + u^\mathcal{A}_i.
\end{align}
Here, \(\gamma_p\) governs the balance between immediate and historical considerations, with \(\gamma_p=0\) reducing to instantaneous fairness and \(\gamma_p=1\) recovering perfect-recall fairness.

\begin{definition}[Discounted-Recall Fairness (Additive)]\label{def:discounted_additive_fairness}
An allocation exhibits \emph{discounted-recall fairness (additive)} at time \(t\) if fairness is evaluated using the past-discounted update. For any candidate allocation \(\mathcal{A}\), define
\begin{align}
    Z^t_i \mid \mathcal{A} = \gamma_p\, Z^{t-1}_i + u^\mathcal{A}_i \quad \text{for } \gamma_p \in [0,1].
\end{align}
A discounted-recall-fair (additive) allocation at time \(t\) is any optimizer
that
selects the allocation that maximizes the welfare applied to the past-discounted cumulative utilities.
\end{definition}

\subsubsection{Discounted Recall with Averaged Utilities}

For averaged utilities, both the accumulated utility and the effective time denominator are discounted. Let \(d_t\) denote the past-discounted denominator at time \(t\). Then:
\begin{align}
    Z^t_i \mid \mathcal{A} = \frac{\gamma_p Z^{t-1}_i \cdot d_{t-1} + u^\mathcal{A}_i}{\gamma_p d_{t-1} + 1}.
\end{align}

\begin{definition}[Discounted-Recall Fairness (Averaged)]\label{def:discounted_averaged_fairness}
An allocation exhibits \emph{discounted-recall fairness (averaged)} at time \(t\) if fairness is evaluated using the past-discounted average with denominator \(d_t\). For any candidate allocation \(\mathcal{A}\), define
\begin{align}
    Z^t_i \mid \mathcal{A} = \frac{\gamma_p Z^{t-1}_i \cdot d_{t-1} + u^\mathcal{A}_i}{\gamma_p d_{t-1} + 1}.
\end{align}
A discounted-recall-fair (averaged) allocation at time \(t\) is any optimizer
that
selects the allocation that maximizes the welfare applied to the past-discounted average utilities.
\end{definition}

In both formulations, the computation of \(\Z^t\) depends only on the previous state—specifically, the augmented state comprising \(\Z^{t-1}\) (and \(d_{t-1}\) in the averaged case). This Markovian structure not only simplifies the computation but also provides a smooth interpolation between instantaneous and perfect-recall fairness, while ensuring that the state space is augmented in a tractable manner.

\subsection{Comparison between the Different Paradigms}

Figure~\ref{fig:combined} illustrates the evolution of the cumulative utility difference, 
$\sum U_{Alice} - \sum U_{Bob},$
under the three fairness approaches: Instantaneous, perfect-recall, and discounted-recall fairness, as discussed in Examples~\ref{ex:inst_example} and \ref{ex:recall_example}. The allocations are made with the egalitarian objective, using additive aggregation for perfect-recall and discounted-recall. 
In Figure~\ref{fig:combined}(left), where both agents participate from the start, instantaneous fairness accumulates short-term differences leading to long-term unfairness, while perfect-recall fairness compensates by accounting for all past allocations. Discounted-recall fairness offers a tunable middle ground, with the discount factor \(\gamma_p\) controlling how quickly past utilities decay in importance.

In Figure~\ref{fig:combined}(center), only Alice is active initially and Bob joins later. All approaches perform similarly in the initial phase. Instantaneous fairness keeps accumulating the imbalance regardless of the history, while both perfect-recall and discounted-recall mechanisms move towards equalizing the past imbalance. Perfect recall only starts allocating resources to Alice again after equalizing the total utility. $\gamma_p$ serves as a tuning knob which lets us control how strongly we want the distant past to affect current allocations.

Finally, Figure~\ref{fig:combined}(right) shows the inner workings of each fairness approach by plotting the perceived differences between Alice and Bob in terms of $\Z$. Instantaneous fairness seems to always keep a low difference between the two agents locally, even as the total utility for Alice keeps rising. Perfect-recall keeps an exact track of resources, considering the distribution unfair even after many steps of allocating to Bob only. The decay of the line, particularly visible for $\gamma_p=0.9$ shows how discounted-recall slowly forgets past decisions, with values close to 1 emulating longer memory.

\section{Theoretical Results}
So far, we have established qualitative and behavioral motivations for the three approaches through illustrative examples.
In this section, we present theoretical results that highlight the practical advantages of past-discounted fairness over perfect-recall fairness. We first focus on the boundedness of the augmented state space, which is crucial for the convergence and scalability of reinforcement learning algorithms, especially in long-horizon settings.

\subsection{Additive Utilities}
A key practical advantage of our past-discounted approach is that it bounds the cumulative utility over time. In typical non-discounted fairness frameworks (e.g., \cite{mcilrathNonMarkovianFairness, zimmer2021MOMDP}), the cumulative utility is computed as
$\Z_i^t = \sum_{\tau=0}^t u_i^\tau,$
which grows linearly with the time horizon (i.e., \(\Z_i^t \le (t+1) u_{\max}\) when \(u_i^t \in [0,u_{\max}]\)). As a consequence, when the fairness state is augmented with these cumulative utilities, the state space expands unboundedly with time, severely hampering the scalability and learnability of the problem.

\begin{theorem}[State explosion under perfect recall]\label{thm:state_explosion_pr}
    Let $u_i^t\in[0,u_{\max}]$ for all agents $i\in\{1,\ldots,n\}$ and times $t\ge 0$, and define perfect-recall memory by
    \begin{align}
        Z_i^t \;=\; \sum_{\tau=0}^{t} u_i^\tau .
    \end{align}
    Then, for any uniform bin width $\Delta>0$, the number of bins per coordinate needed to represent $Z_i^t$ on a uniform grid satisfies
    \begin{align}
        N_{\mathrm{bins}}^{\mathrm{PR}}(t,\Delta) \;\ge\; \Big\lceil \frac{(t+1)\,u_{\max}}{\Delta} \Big\rceil ,
    \end{align}
    so the grid over $\Z^t\in\mathbb{R}^n$ has cardinality at least $\big(N_{\mathrm{bins}}^{\mathrm{PR}}(t,\Delta)\big)^{n}$, which grows linearly with $t$ per coordinate and exponentially with $n$. The linear dependence on $t$ is tight.
\end{theorem}

\begin{proof}
    For each $i$,
    \begin{align}
        0 \;\le\; Z_i^t \;=\; \sum_{\tau=0}^{t} u_i^\tau \;\le\; \sum_{\tau=0}^{t} u_{\max} \;=\; (t+1)\,u_{\max}.
    \end{align}
    A uniform grid with bin width $\Delta$ covering $[0,(t+1)u_{\max}]$ requires at least $\big\lceil (t+1)u_{\max}/\Delta \big\rceil$ bins per coordinate. Across $n$ coordinates (agents), the number of grid cells multiplies, yielding the stated growth. Tightness holds by taking $u_i^\tau\equiv u_{\max}$, for which $Z_i^t=(t+1)u_{\max}$.
\end{proof}

\begin{theorem}[Horizon-independent boundedness under past discounting]\label{thm:bounded_pd}
    Fix $\gamma_p\in[0,1)$ and $u_{\max}>0$. Let $u_i^t\in[0,u_{\max}]$ and define the past-discounted memory by
    \begin{align}
        Z_i^t \;=\; \gamma_p\, Z_i^{t-1} + u_i^t, \qquad Z_i^0 \;=\; u_i^0 .
    \end{align}
    Then, for all $t\ge 0$,
    \begin{align}
        0 \;\le\; Z_i^t \;\le\; \frac{u_{\max}}{1-\gamma_p}
        \qquad\text{and}\qquad
        Z_i^t \;=\; \sum_{k=0}^{t} \gamma_p^{\,t-k} u_i^k .
    \end{align}
    The upper bound is tight: if $u_i^t\equiv u_{\max}$, then $Z_i^t = u_{\max}/(1-\gamma_p)$.
\end{theorem}

\begin{proof}
    Unrolling the recursion yields
    \begin{align}
        Z_i^t \;=\; \sum_{k=0}^{t} \gamma_p^{\,t-k} u_i^k .
    \end{align}
    Since $u_i^k\in[0,u_{\max}]$ and $\gamma_p^{\,t-k}\ge 0$,
    \begin{align}
        0 \;\le\; Z_i^t \;\le\; u_{\max} \sum_{k=0}^{t} \gamma_p^{\,t-k}
        \;=\; u_{\max} \sum_{j=0}^{t} \gamma_p^{\,j}
        \;\le\; \frac{u_{\max}}{1-\gamma_p}.
    \end{align}
    Moreover, the bound is tight. If $u_i^t \equiv u_{\max}$, then
    \begin{align}
        Z_i^t \;=\; u_{\max}\sum_{j=0}^{t}\gamma_p^{\,j}
        \;=\; \frac{u_{\max}}{1-\gamma_p}\,\bigl(1-\gamma_p^{\,t+1}\bigr),
    \end{align}
    so $Z_i^t$ is increasing in $t$ and 
    \begin{align}
        \lim_{t\to\infty} Z_i^t \;=\; \frac{u_{\max}}{1-\gamma_p}.
    \end{align}
    
\end{proof}

\begin{corollary}[Horizon-independent discretization under past discounting]\label{cor:discretization_pd}
    Under the conditions of Theorem~\ref{thm:bounded_pd}, any uniform grid with bin width $\Delta>0$ requires at most
    \begin{align}
        N_{\mathrm{bins}}^{\mathrm{PD}}(\Delta) \;=\; \Big\lceil \frac{u_{\max}}{(1-\gamma_p)\,\Delta} \Big\rceil
    \end{align}
    bins per coordinate—independent of $t$. Consequently, a fixed-resolution discretization of the augmented fairness state $\Z^t$ has size independent of the horizon; across $n$ coordinates the grid has at most $\big(N_{\mathrm{bins}}^{\mathrm{PD}}(\Delta)\big)^{n}$ cells.
\end{corollary}

\bigskip

\subsection{Averaged Utilities}
Theorems~\ref{thm:state_explosion_pr} and \ref{thm:bounded_pd} together show that past-discounted fairness avoids the state explosion inherent in perfect-recall fairness for additive aggregation. Next we provide similar results for the averaged aggregation.

\begin{theorem}[Averaged perfect recall requires unbounded augmented state]\label{thm:avg_pr_unbounded}
Define the running average update by
\begin{align}
  Z_i^{t} \;=\; \frac{(t-1)\,Z_i^{t-1} + u_i^{t}}{t}, \qquad Z_i^{0}=u_i^{0},
\end{align}
with $u_i^{t}\in[0,u_{\max}]$. Then:
\begin{itemize}
\item[(i)] (\emph{Non-Markov in $\Z^t$ alone}) There exist histories $h,h'$ with the same $\Z^t$ but different $t$ such that, for the same $u_i^{t+1}$, the next value $Z_i^{t+1}$ differs. Hence the process is not Markov in $(s^t,\Z^t)$ without carrying $t$ (or an equivalent denominator).
\item[(ii)] (\emph{Unbounded augmentation}) Any Markov augmentation that appends $t$ yields an auxiliary variable whose support is $\{0,1,2,\dots\}$, i.e., unbounded in the horizon.
\end{itemize}
\end{theorem}

\begin{proof}
(i) From the update,
\begin{align}
  Z_i^{t+1} \;=\; \frac{t\,Z_i^{t} + u_i^{t+1}}{t+1},
\end{align}
which depends on $t$ even when $Z_i^{t}$ and $u_i^{t+1}$ are fixed. Take two histories with the same $Z_i^t=z$ but different $t$ and the same $u_i^{t+1}$; the resulting $Z_i^{t+1}$ differ, so $\Z^t$ alone is insufficient for Markov evolution.  
(ii) Appending $t$ (or any one-to-one proxy such as the exact denominator) restores Markovian evolution but makes the augmented state include a variable that grows without bound as $t\to\infty$.
\end{proof}

\begin{theorem}[Past-discounted averaging admits a bounded Markov augmentation]\label{thm:avg_pd_bounded}
Let $\gamma_p\in[0,1)$ and define $d_0=0$, $d_{t}=\gamma_p d_{t-1}+1$, and
\begin{align}
  Z_i^{t} \;=\; \frac{\gamma_p d_{t-1}\, Z_i^{t-1} + u_i^{t}}{\gamma_p d_{t-1}+1}, \qquad u_i^{t}\in[0,u_{\max}].
\end{align}
Then:
\begin{itemize}
\item[(i)] (\emph{Boundedness}) $d_t = \sum_{k=0}^{t-1}\gamma_p^{\,k} \le \tfrac{1}{1-\gamma_p}$ and $0\le Z_i^{t}\le u_{\max}$ for all $t$.
\item[(ii)] (\emph{Markov sufficiency}) The pair $(\Z^t,d_t)$ is a sufficient memory: the next $(\Z^{t+1},d_{t+1})$ depends only on $(\Z^t,d_t)$ and current utilities.
\item[(iii)] (\emph{Non-vanishing responsiveness}) The marginal weight of the current utility in $Z_i^{t}$ equals
\begin{align}
  \frac{\partial Z_i^{t}}{\partial u_i^{t}} \;=\; \frac{1}{\gamma_p d_{t-1}+1}
  \;\xrightarrow[t\to\infty]{}\; 1-\gamma_p,
\end{align}
so the influence of the current step does not vanish (contrast with the $1/t$ weight under plain averaging).
\end{itemize}
\end{theorem}

\begin{proof}
(i) The recursion for $d_t$ solves to $d_t=\sum_{k=0}^{t-1}\gamma_p^{\,k}\le\frac{1}{1-\gamma_p}$. Multiplying the $Z$-update by $\gamma_p d_{t-1}+1$ gives
\begin{align}
  (\gamma_p d_{t-1}+1)\, Z_i^{t} \;=\; \gamma_p d_{t-1}\, Z_i^{t-1} + u_i^{t},
\end{align}
so $Z_i^{t}$ is a convex combination of $Z_i^{t-1}$ and $u_i^{t}$ with weights summing to $1$; by induction and $u_i^{t}\in[0,u_{\max}]$, we get $Z_i^{t}\in[0,u_{\max}]$.
(ii) $(\Z^t,d_t)$ updates deterministically from $(\Z^t,d_t)$ and $\mathbf u^{t}$ via the two recursions, hence is Markov-sufficient. Both coordinates are bounded by part (i).
(iii) Differentiate the update with respect to $u_i^{t}$; the coefficient is $(\gamma_p d_{t-1}+1)^{-1}$. Using $d_{t-1}\to \tfrac{1}{1-\gamma_p}$ yields the limit $1-\gamma_p$.
\end{proof}

\xhdr{Implications:}
Taken together, these results isolate why past discounting is preferable for long-horizon learning. With additive \emph{perfect recall} (Theorem~\ref{thm:state_explosion_pr}), the augmented fairness state necessarily grows with the horizon; any fixed-resolution representation explodes linearly in \(t\) per coordinate (and exponentially in \(n\)). With \emph{averaged perfect recall} (Theorem~\ref{thm:avg_pr_unbounded}), the values themselves stay bounded, but Markovization forces carrying an unbounded auxiliary variable (the time index or equivalent denominator), so the augmentation still scales with the horizon. In contrast, \emph{past-discounted} schemes yield horizon-independent summaries: additive discounting gives a tight uniform bound on \(Z_i^t\) (Theorem~\ref{thm:bounded_pd}) and the number of additional states introduced by a fixed discretization is independent of the time horizon (Corollary~\ref{cor:discretization_pd}). Discounted averaging also provides a bounded sufficient statistic \((\Z^t,d_t)\) with non-vanishing responsiveness of the current step (Theorem~\ref{thm:avg_pd_bounded}). Practically, this means that even in long-horizon settings, the augmented state space remains fixed in size, enabling efficient learning and planning. Moreover, the non-vanishing responsiveness ensures that recent allocations meaningfully influence fairness evaluations, preventing the system from becoming overly rigid due to distant past decisions.

\subsection{Half-life and Effective Memory}

With past discounting, we also want to decide how much of the past to remember. The discount factor \(\gamma_p\) controls this trade-off: values close to 1 retain more history, while values near 0 emphasize recent allocations. However, we still need a systematic way to choose \(\gamma_p\) based on the desired memory characteristics.

We introduce the term \emph{half-life} to quantify how quickly past outcomes lose influence under past discounting. Under the additive update \(Z_i^t=\gamma_p Z_i^{t-1}+u_i^t\), the contribution of a utility observed \(k\) steps ago is weighted by \(\gamma_p^{\,k}\) at time \(t\). The \emph{half-life} \(t_{1/2}\) is the number of steps after which this weight falls to one half of its initial value, i.e., it is defined by \(\gamma_p^{\,t_{1/2}}=\tfrac{1}{2}\). This provides an interpretable way to choose \(\gamma_p\) based on how long past allocations should meaningfully affect current fairness.

\begin{definition}[Half-life]\label{def:half_life}
Under the additive past-discounted update $Z_i^t=\gamma_p Z_i^{t-1}+u_i^t$ with $\gamma_p\in(0,1)$, the \emph{half-life} $t_{1/2}$ is the number of steps after which the weight on a past contribution is halved, i.e.
\begin{align}
    \gamma_p^{\,t_{1/2}}=\tfrac12
    \qquad\Longleftrightarrow\qquad
    t_{1/2}=\frac{\ln(1/2)}{\ln(\gamma_p)}.
\end{align}
\end{definition}

\xhdr{Effective window (how much history is retained in total).}
The geometric weights \((1,\gamma_p,\gamma_p^2,\ldots)\) have total mass
\begin{align}
  \sum_{k=0}^{\infty}\gamma_p^{k}=\frac{1}{1-\gamma_p}.
\end{align}
It is convenient to call 
\begin{align}
  W_{\mathrm{eff}} := \frac{1}{1-\gamma_p}
\end{align}
the \emph{effective window length}. Think of it as: “discounting with \(\gamma_p\) behaves, in total weight, like remembering roughly \(W_{\mathrm{eff}}\) recent steps equally.” Two immediate consequences:
\begin{align}
  \gamma_p = 1-\frac{1}{W_{\mathrm{eff}}}
  \qquad\text{and}\qquad
  \frac{\partial Z_i^t}{\partial u_i^t}=1-\gamma_p=\frac{1}{W_{\mathrm{eff}}}.
\end{align}
So \(W_{\mathrm{eff}}\) directly tells you the \emph{instantaneous weight} on the current step (larger window \(\Rightarrow\) smaller immediate weight, more smoothing).

\xhdr{``How much of the last \(m\) steps should matter?''}
If you want at least a fraction \(1-\varepsilon\) of the total influence to come from the most recent \(m\) steps, choose \(\gamma_p\) so that
\begin{align}
  \underbrace{\frac{\sum_{k=0}^{m-1}\gamma_p^{k}}{\sum_{k=0}^{\infty}\gamma_p^{k}}}_{\text{fraction from last $m$}} 
  = 1-\gamma_p^{m} \;\ge\; 1-\varepsilon
  \quad\Longleftrightarrow\quad
  \gamma_p \;\le\; \varepsilon^{1/m}.
\end{align}
Equivalently, for a fixed \(\gamma_p\), the last \(m\) steps account for \(1-\gamma_p^{m}\) of the total weight.


Table~\ref{tab:half_life_examples} gives half-life and effective window for common \(\gamma_p\).

\begin{table}[h!]
\centering
\caption{Half-life and effective window for selected \(\gamma_p\).}
\begin{tabular}{rcc}
\toprule
\(\gamma_p\) & \(t_{1/2}=\ln(1/2)/\ln(\gamma_p)\) & \(W_{\mathrm{eff}}\approx 1/(1-\gamma_p)\) \\
\midrule
0.80 & 3.11 & 5 \\
0.90 & 6.58 & 10 \\
0.95 & 13.52 & 20 \\
0.97 & 22.77 & 33.3 \\
0.99 & 68.97 & 100 \\
\bottomrule
\end{tabular}
\label{tab:half_life_examples}
\end{table}

\subsection{Beyond Scalar Discounting: Linear Fairness Memories}

We also connect our results to the well-established notion of system stability in control theory. While we show past discounts as one surefire method of keeping the augmented state-space bounded, other mechanisms may exist that achieve similar outcomes. 
We propose this extension
by modeling the fairness memory as a Linear Time-Invariant (LTI) system. This powerful and well-established framework allows us to derive a universal condition for tractable memory updates. We represent the evolution of the n-agent memory vector $\mathbf{Z}^t \in \mathbb{R}^n$ with the following state-space equation:
\begin{align}
    \mathbf{Z}^{t+1} = A\mathbf{Z}^{t} + \mathbf{u}^{t+1}
\end{align}
Here, $\mathbf{u}^{t+1}$ is the vector of utilities received at the current step, and the matrix $A \in \mathbb{R}^{n \times n}$ is the \textbf{memory transition matrix}. This matrix $A$ encodes how the entire memory vector from the previous step is transformed and carried forward, generalizing additive utility aggregation. The central question becomes: for which matrices $A$ will the memory $\mathbf{Z}^t$ remain bounded over arbitrarily long horizons?



The answer is provided by a cornerstone result from control theory, which we state here for completeness.

\begin{theorem}[LTI System Stability]\label{thm:lti_stability}
Consider the update $\mathbf{Z}^{t+1}=A\mathbf{Z}^{t}+\mathbf{u}^{t+1}$ with $A\in\mathbb{R}^{n\times n}$ and $\|\mathbf{u}^{t}\|_\infty\le u_{\max}$ for all $t$. The state vector $\mathbf{Z}^{t}$ is uniformly bounded for all bounded input sequences $\{\mathbf{u}^{t}\}$ if and only if the spectral radius of A is less than one, i.e., $\rho(A) < 1$.
\end{theorem}
\begin{proof}
This is a classic result from linear systems theory establishing the condition for Bounded-Input, Bounded-State (BIBS) stability. For a formal proof, see, e.g., \cite{ogata1995discrete}.
\end{proof}

\xhdr{Implications for Temporal Fairness.}
This general theorem provides a powerful lens through which to view our previous findings, grounded in fundamental principles of system stability.
\begin{itemize}
    \item \textbf{Perfect Recall}, where $A = I$ (the identity matrix), has a spectral radius of $\rho(I) = 1$. The theorem correctly predicts that this system is unstable and its memory will grow unboundedly, formalizing the core problem we identified in Theorem~\ref{thm:state_explosion_pr}.

    \item \textbf{Additive Past-Discounting}, where $A = \gamma_p I$, has a spectral radius of $\rho(\gamma_p I) = |\gamma_p|$. As long as $\gamma_p < 1$, the stability condition is met, guaranteeing a bounded memory. This shows that our central result in Theorem~\ref{thm:bounded_pd} is also supported by this general stability principle, lending further credence to the use of past-discounts.
\end{itemize}

This reveals that any attempt to incorporate long-term memory into fairness must satisfy the condition $\rho(A) < 1$.
More generally, the connection of additive fairness memories to Theorem~\ref{thm:lti_stability} outlines a design rule for richer memory paradigms that allow cross-agent mixing or modeling memory across multiple time scales as long as $A$ is Schur-stable. 
This provides a rigorous foundation for extending our work. While our method is demonstrably effective, this framework offers a clear path to design more sophisticated and nuanced fairness memories while retaining tractability.

\section{Experimental Results}

To empirically validate our theoretical findings, we design an experiment to demonstrate the computational necessity of past-discounting for learning fair policies in long-horizon settings. Our central hypothesis is that with longer horizons a reinforcement learning (RL) agent optimizing for long-term fairness will fail to learn effectively under perfect recall due to the state-space explosion predicted by our theory, while an agent using past-discounting will learn successfully and tractably.

\subsection{Experimental Setup}


\xhdr{Environment.} We use a dynamic resource allocation environment with $n=10$ agents over an episode of length $T$. At each timestep, $n_r=2$ indivisible resources become available. Each agent $i$ has an immediate utility for the resources, represented by a ``needs'' vector, drawn from $\mathcal{U}(0,1)$. If allocated a resource, the agent's immediate utility is equal to this value. To create a challenging fairness scenario, we designate two ``advantaged agents'' whose needs are consistently drawn from a higher distribution, $\mathcal{U}(0.8,1.0)$. This creates a natural tension where a myopic, efficiency-maximizing policy would perpetually favor the advantaged agents, leading to severe long-term inequality.

\xhdr{Learning Problem.} We frame the task as an RL problem where a central allocator must learn a policy $\pi(s_t) \to a_t$.
\begin{itemize}
    \item \textbf{State ($s_t$):} The state consists of the current needs vector and the fairness memory vector from the previous step, $\mathbf{Z}^{t-1}$.
    \item \textbf{Action ($a_t$):} The action is a discrete choice from the set of all $\binom{n}{n_r}$ possible pairs of agents to receive the resources.
    \item \textbf{Reward ($r_t$):} The reward signal is a weighted sum of an efficiency component $U_t$ and a potential-based fairness component $F_t$:
    $r_t = (1-\lambda) \cdot U_t + \lambda \cdot F_t$.\\
    In our experiments, we set $\lambda=0.9$, as our main objective is to learn to improve fairness, with utility acting as a secondary shaping reward. The efficiency term $U_t = \sum_{i} u_i^t$ is the total utility allocated at the current step. The fairness term $F_t = W(\mathbf{Z}^t) - W(\mathbf{Z}^{t-1})$ is the change in the social welfare function, which directly rewards actions that improve the fairness state.
\end{itemize}
The agent's goal is to maximize the cumulative reward. We conduct experiments using two distinct welfare functions, $W(\cdot)$: \textbf{Egalitarian Welfare} and \textbf{Nash Welfare}.


\xhdr{Models.} We train a Proximal Policy Optimization (PPO) agent~\cite{schulman2017PPO} from the Stable Baselines3 library~\cite{raffin2021stable} for 1 million total timesteps under three different memory configurations, corresponding directly to the paradigms discussed in our paper:
\begin{enumerate}
    \item \textbf{Perfect-Recall ($\gamma_p=1.0$):} The agent observes the true cumulative utility, $\mathbf{Z}^t = \sum_{\tau=0}^t \mathbf{u}^\tau$. As predicted by Theorem~\ref{thm:state_explosion_pr}, this leads to an unbounded state space.
    \item \textbf{Past-Discounted ($\gamma_p\in\{0.9,0.99,0.999\}$):} The agent observes the discounted utility, $\mathbf{Z}^t = \gamma_p \mathbf{Z}^{t-1} + \mathbf{u}^t$. Following Theorem~\ref{thm:bounded_pd}, this ensures the state space remains bounded.
    \item \textbf{Myopic ($\gamma_p=0.0$):} The agent only observes immediate needs, lacking any historical context to correct for long-term imbalances.
\end{enumerate}

We conduct our experiments across different horizons ranging from 100 steps per episode to 10000 steps per episode. Each configuration was run 10 times for statistical confidence. Further details are provided in the supplement.

\begin{figure}[t]
    \centering
    \includegraphics[width=0.72\linewidth]{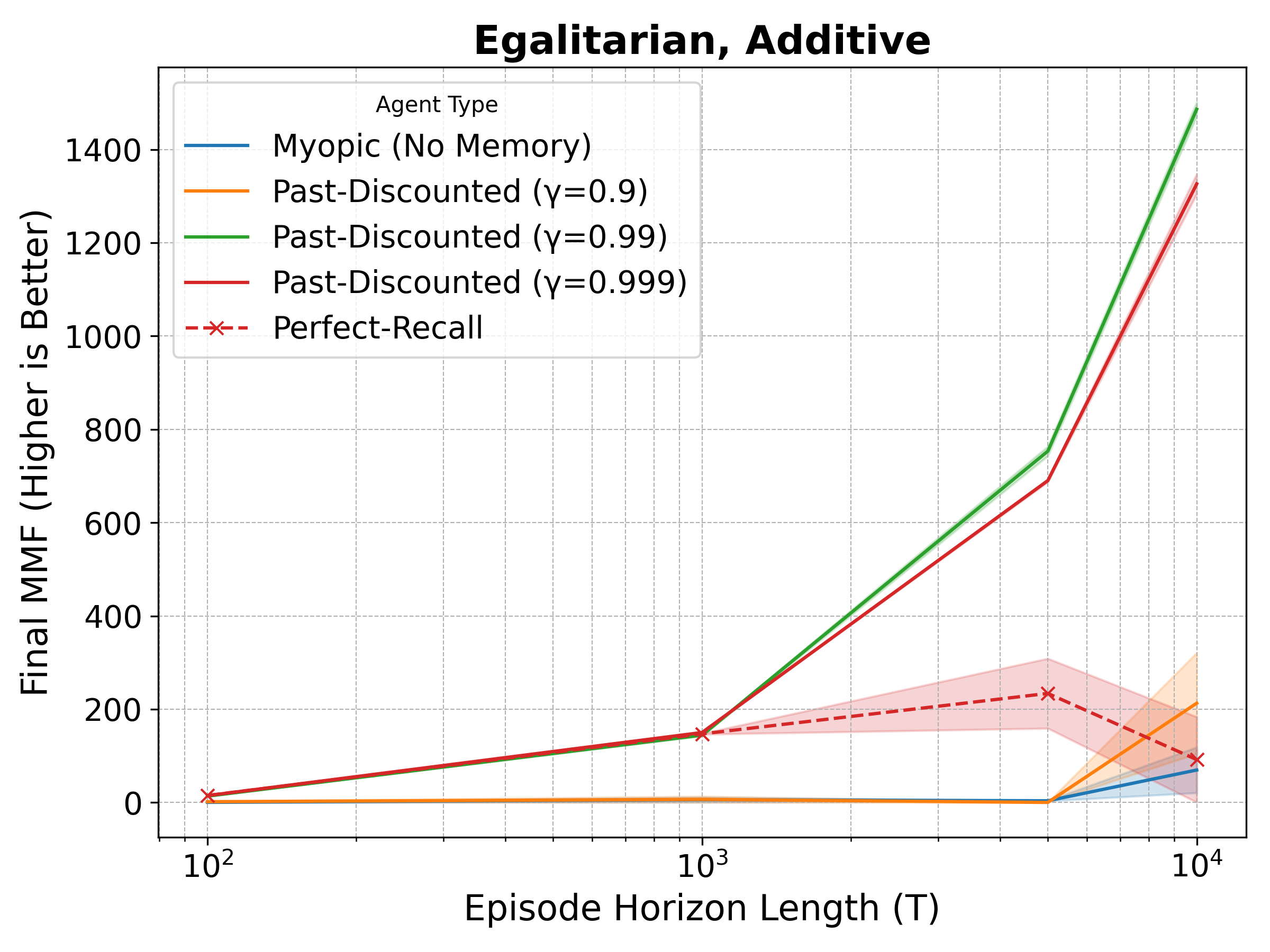}
    \includegraphics[width=0.72\linewidth]{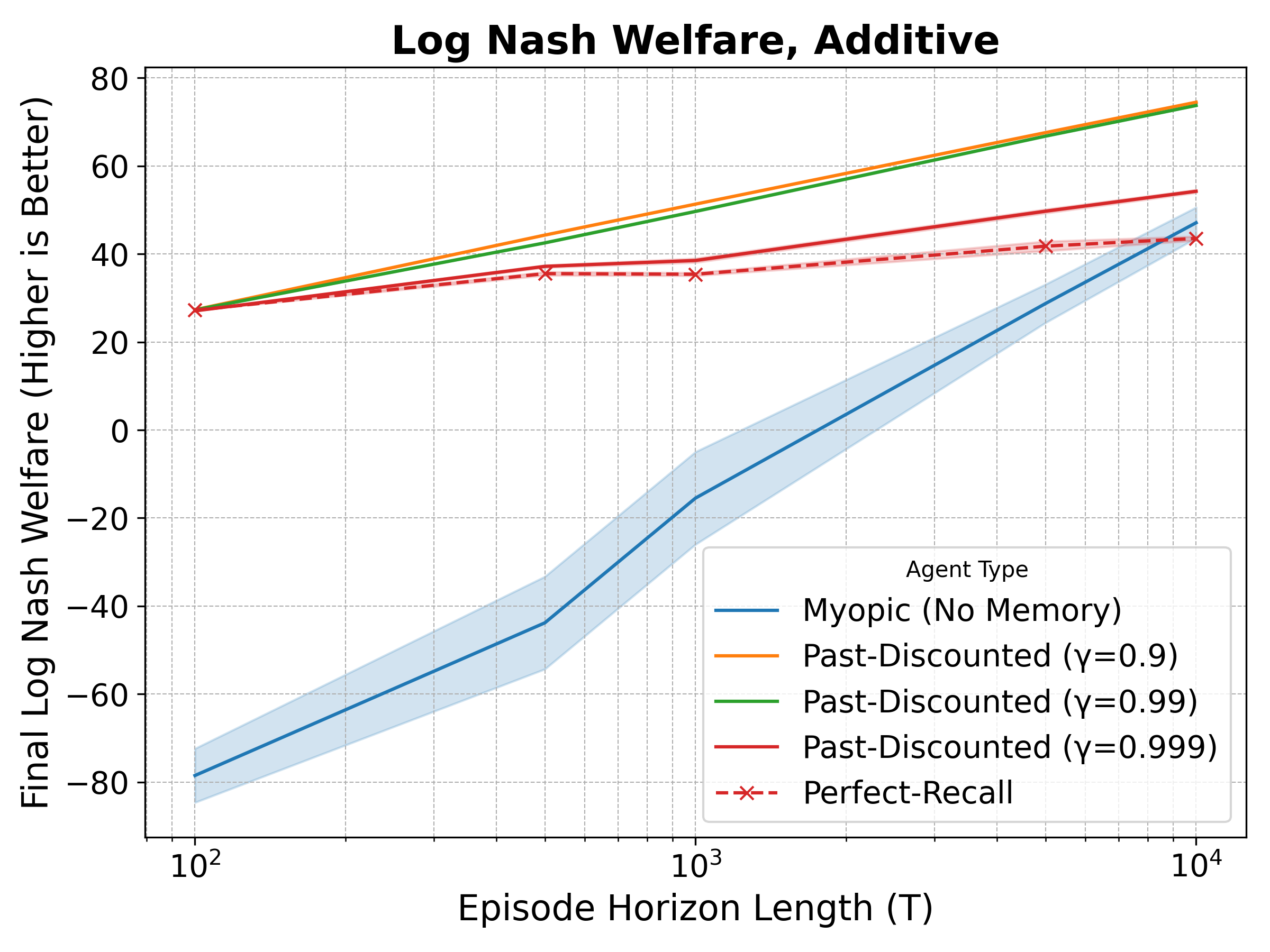}
    \caption{Final converged welfare function value as the episode length is increased. Perfect Recall sees a degradation in performance as horizon is increased. 
    }
    \label{fig:exp_results}
\end{figure}

\subsection{Results}

\begin{figure}[t]
    \centering
    \includegraphics[width=0.72\linewidth]{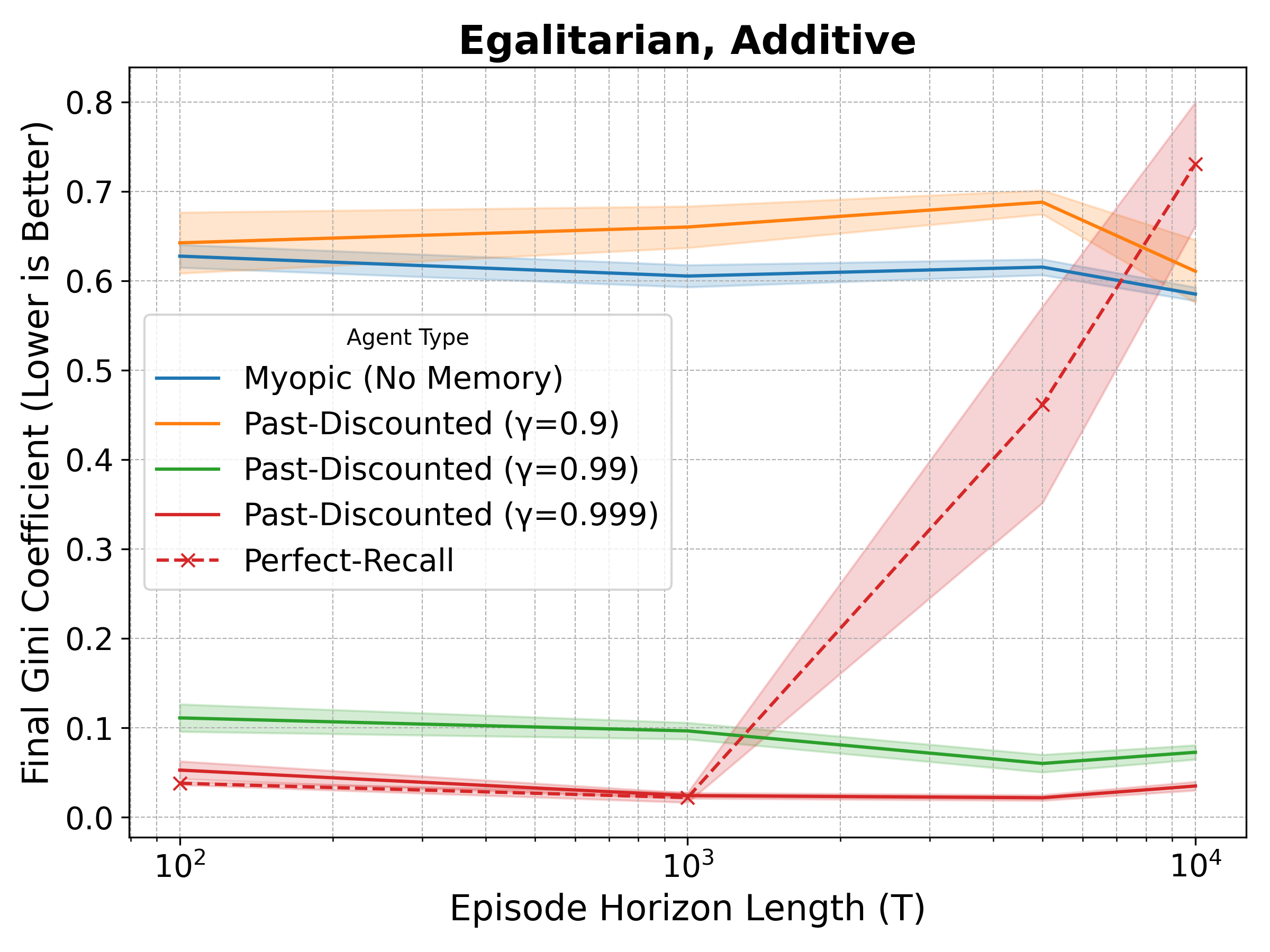}
    \includegraphics[width=0.72\linewidth]{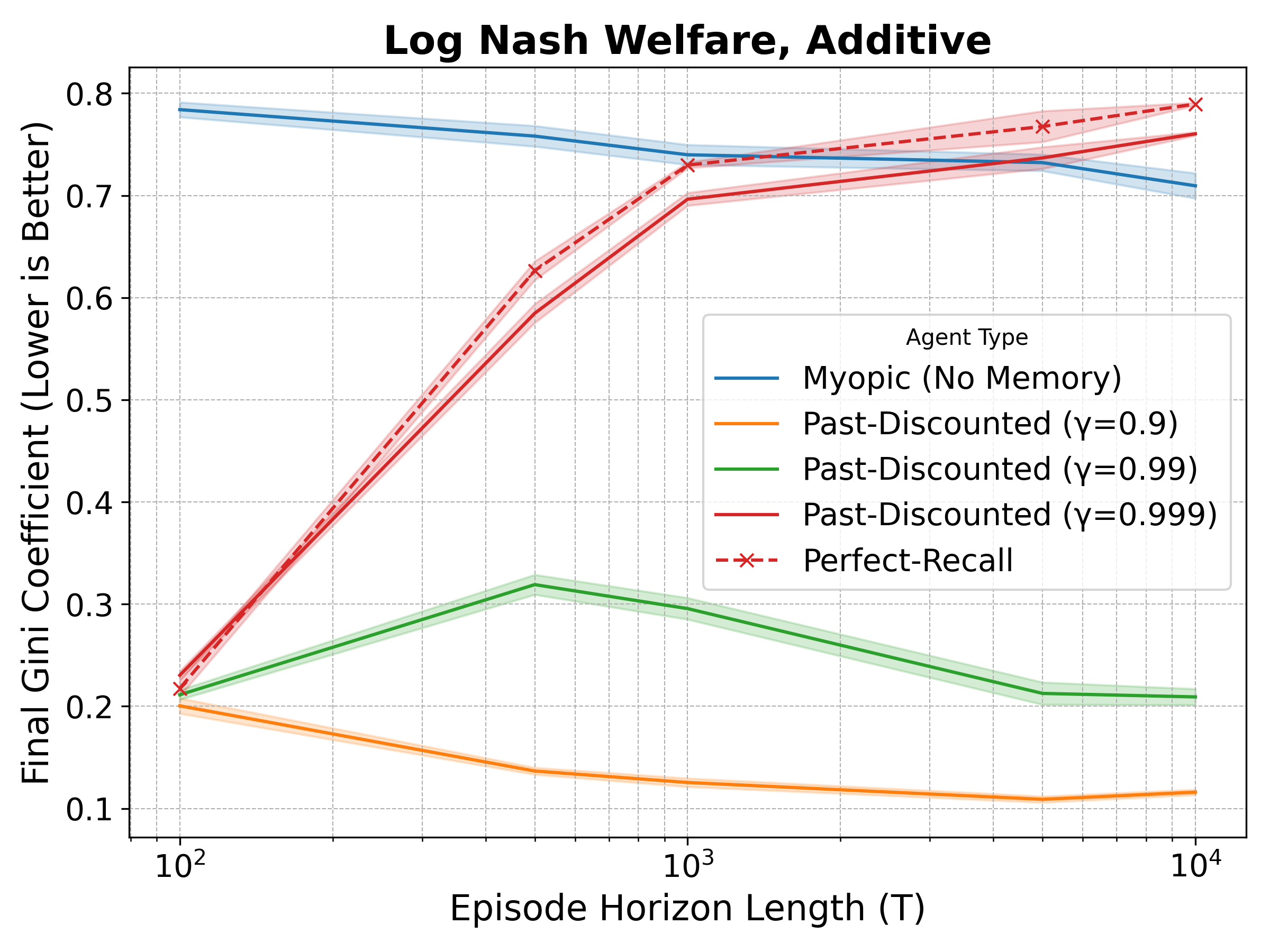}
    \caption{Gini coefficient of the converged distribution as the episode length is increased. Perfect Recall sees a degradation in performance as horizon is increased.}
    \label{fig:exp_results_gini}
\end{figure}

We compare the converged performance of each method after training for 1 million steps, computing metrics by averaging over the last 10\% of the episodes. 
We additionally evaluate the performance of each agent using the Gini coefficient, a standard measure of inequality where 0 represents perfect equality. The results are shown in Figures~\ref{fig:exp_results} and~\ref{fig:exp_results_gini}.

The \textbf{Myopic} agent performs poorly across all scenarios, confirming that memory is essential for learning fair behavior. By only observing immediate needs, it cannot correct for the accumulating advantage of certain agents, resulting in high inequality.


The \textbf{Perfect-Recall} agent demonstrates the critical failure mode predicted by Theorem~\ref{thm:state_explosion_pr}. While effective at short horizons ($T \le 500$), its performance collapses as the episode length grows. At $T=10,000$, its ability to maintain fairness is no better than the myopic agent in both experiments. This failure is a direct consequence of its unbounded state space; as the cumulative utilities in $\mathbf{Z}^t$ grow, it makes learning harder, preventing the PPO algorithm from converging to a stable and effective policy.

On the other hand, \textbf{Past-Discounted} agents are able to learn regardless of the horizon. Benefiting from a stable, bounded state representation, they can learn a robust policy that maintains a low Gini coefficient even over this extended horizon. 

Diving deeper into the Past-Discounted results reveals how the choice of the discount factor, $\gamma_p$, is critical for performance. The agent with $\gamma_p=0.999$, which has an effective memory window of 1000 steps, closely mimics the behavior of the Perfect-Recall agent on horizons up to that length. Beyond a 1000-step horizon, however, the methods diverge: the Perfect-Recall agent's performance collapses, while the Past-Discounted agent's performance improves.

Furthermore, our results show that the optimal choice of $\gamma_p$ depends on the specific fairness objective. With the dense signal from \textbf{Nash Welfare}, a shorter memory ($\gamma_p=0.9, W_{eff}=10$) is sufficient for the agent to learn a fair policy. However, with the sparser signal from \textbf{Egalitarian Welfare}—where the reward only changes when the identity of the worst-off agent is affected—a longer memory ($\gamma_p=0.99, W_{eff}=100$) is required to learn effectively.

This highlights a key insight for practitioners: selecting an appropriate discount factor involves balancing the complexity of the environment with the nature of the fairness signal. Denser signals may allow for shorter effective memories, while sparser signals necessitate longer ones ($\text{higher } \gamma_p$) to ensure the agent accumulates enough information to act fairly. In our experiments, a $\gamma_p$ of $0.99$ proved to be a robust choice across both welfare functions.

These results provide compelling empirical validation for our central thesis. While perfect recall is theoretically appealing, its unbounded memory requirement makes it computationally intractable for learning fair policies over long horizons. Past-discounting is not merely a heuristic but a practical necessity, providing a bounded and stationary state representation that enables stable learning and scalable long-term fairness.



\section{Conclusion}

In this work, we introduced a framework for incorporating past-discounted historical utilities into dynamic resource allocation—a strategy inspired by behavioral economics and moral psychology, which show that humans naturally discount the impact of distant past events \cite{trope2010construal,frederick2002time}. By applying a discount factor \(\gamma_p\) to past utilities, our method enables decision-makers to reason over accumulated utilities while effectively balancing short-term and long-term fairness considerations.

Crucially, the past-discounting approach ensures that the augmented state space remains bounded. In contrast to traditional non-discounted fairness methods—where the state space grows linearly with the time horizon and thus becomes computationally intractable—our framework yields a joint state space whose size is independent of the time horizon. This boundedness not only improves the sample complexity and convergence of reinforcement learning algorithms in multi-agent settings, but also provides a principled mechanism to manage the trade-off between immediate outcomes and historical context.






\bibliographystyle{ACM-Reference-Format} 
\bibliography{main}


\newpage
\appendix
\input{appendix-contents}
\end{document}

%% file: appendix-contents.tex
\section{Extended Experiment Details}
\label{sec:appendix_extended_details}

This section provides additional implementation details to ensure the reproducibility of our results.

\xhdr{Metric Calculation vs. Reward Signal}
A key methodological detail is the distinction between the reward signal used for training and the metrics reported in our results. The agent’s potential-based fairness reward, $F_t = W(\mathbf{Z}^t) - W(\mathbf{Z}^{t-1})$, was calculated using the specific memory model being tested (e.g., past-discounted $\mathbf{Z}^t$). However, to provide a consistent and unbiased evaluation across all models, the final reported Gini coefficients and cumulative utilities are always calculated using the \textbf{true, undiscounted cumulative utility} for each agent over the entire episode. This ensures that we are comparing the true long-term outcomes of each policy, independent of the agent's internal memory representation.

\xhdr{Log Nash Welfare Implementation}
For experiments using Nash Welfare, we optimized the logarithm of the Nash product. This improves numerical stability by preventing underflow from multiplying many small utility values and transforms the multiplicative objective into an additive one, which is more amenable to learning. To avoid `nan` or `-inf` values from $\log(0)$ in cases where an agent has zero utility, we added a small constant, $\epsilon = 10^{-6}$, to each agent's utility before taking the logarithm. The objective was therefore:
$$ W_{\text{log-Nash}}(\mathbf{Z}) = \sum_{i=1}^{n} \log(z_i + \epsilon) $$

\xhdr{State and Action Spaces}
The agent's observation at each timestep $t$ was a flattened vector of size 20, formed by concatenating the 10-dimensional agent "needs" vector with the 10-dimensional fairness memory vector $\mathbf{Z}^{t-1}$. All values were normalized to the approximate expected range. The action space was discrete with $\binom{10}{2} = 45$ actions, where each action corresponds to selecting a unique pair of agents to receive the two available resources.

\xhdr{Code Availability}
The source code for our environment and experimental runner is included in the supplement attachment.

\begin{table}[ht!]
\centering
\caption{Experimental settings and hyperparameters.}
\label{tab:hyperparams}
\begin{tabular}{ll}
\toprule
\textbf{Hyperparameter} & \textbf{Value} \\
\midrule
\multicolumn{2}{l}{\textit{Reinforcement Learning}} \\
RL Algorithm & Proximal Policy Optimization (PPO) \\
Policy Network & MlpPolicy \\
Total Timesteps & 1,000,000 \\
PPO Hyperparameters & Default Stable Baselines3 values \\
\midrule
\multicolumn{2}{l}{\textit{Environment}} \\
Number of Agents & 10 \\
Resources per Step & 2 \\
Episode Length (Steps) & \{100, 500, 1000, 5000, 10000\} \\
Fairness Weight ($\lambda$) & 0.9 (0.999 for Averaged Egalitarian) \\
\midrule
\multicolumn{2}{l}{\textit{Experimental Setup}} \\
Runs per Configuration & 10 (different seeds) \\
Memory Models Tested & Myopic, Perfect-Recall, Past-Discounted \\
Past-Discount Factors ($\gamma_p$) & \{0.0, 0.9, 0.99, 0.999, 1.0\} \\
Aggregation Types & Additive, Averaged \\
\midrule
\multicolumn{2}{l}{\textit{Hardware}} \\
Machine & Apple Mac Mini (M4, 64GB RAM) \\
Approx. Runtime per Run & $\sim$5 minutes \\
\bottomrule
\end{tabular}
\end{table}
\section{Model Settings and Hyperparameters}
\label{sec:appendix_hyperparams}

All experiments were conducted using the Proximal Policy Optimization (PPO) algorithm from the Stable Baselines3 library. The agent's policy was represented by a Multi-Layer Perceptron (MlpPolicy). For all PPO-specific hyperparameters, such as learning rate, batch size, and clipping parameters, we used the default values provided by the Stable Baselines3 implementation, which are established baselines for reliable performance.

Each experimental configuration (i.e., a specific combination of fairness model, welfare function, and episode length) was trained for a total of 1 million timesteps. To ensure statistical significance, every configuration was run 10 times with different random seeds.

The resource allocation environment, detailed in the main paper, consists of 10 agents and 2 available resources per timestep. The episode length was varied as a key experimental parameter, with values of 100, 500, 1000, 5000, and 10000 steps. The fairness weight in the reward function was set to $\lambda=0.9$ for all experiments, with the exception of the Averaged Egalitarian runs, which used $\lambda=0.999$ to ensure a stable learning signal.

All experiments were executed on a Mac Mini with an M4 processor and 64GB of RAM. A single training run of 1 million timesteps took approximately 5 minutes to complete. A summary of the key parameters is provided in Table~\ref{tab:hyperparams}.

\begin{figure}[h!]
    \centering
    \includegraphics[width=0.48\linewidth]{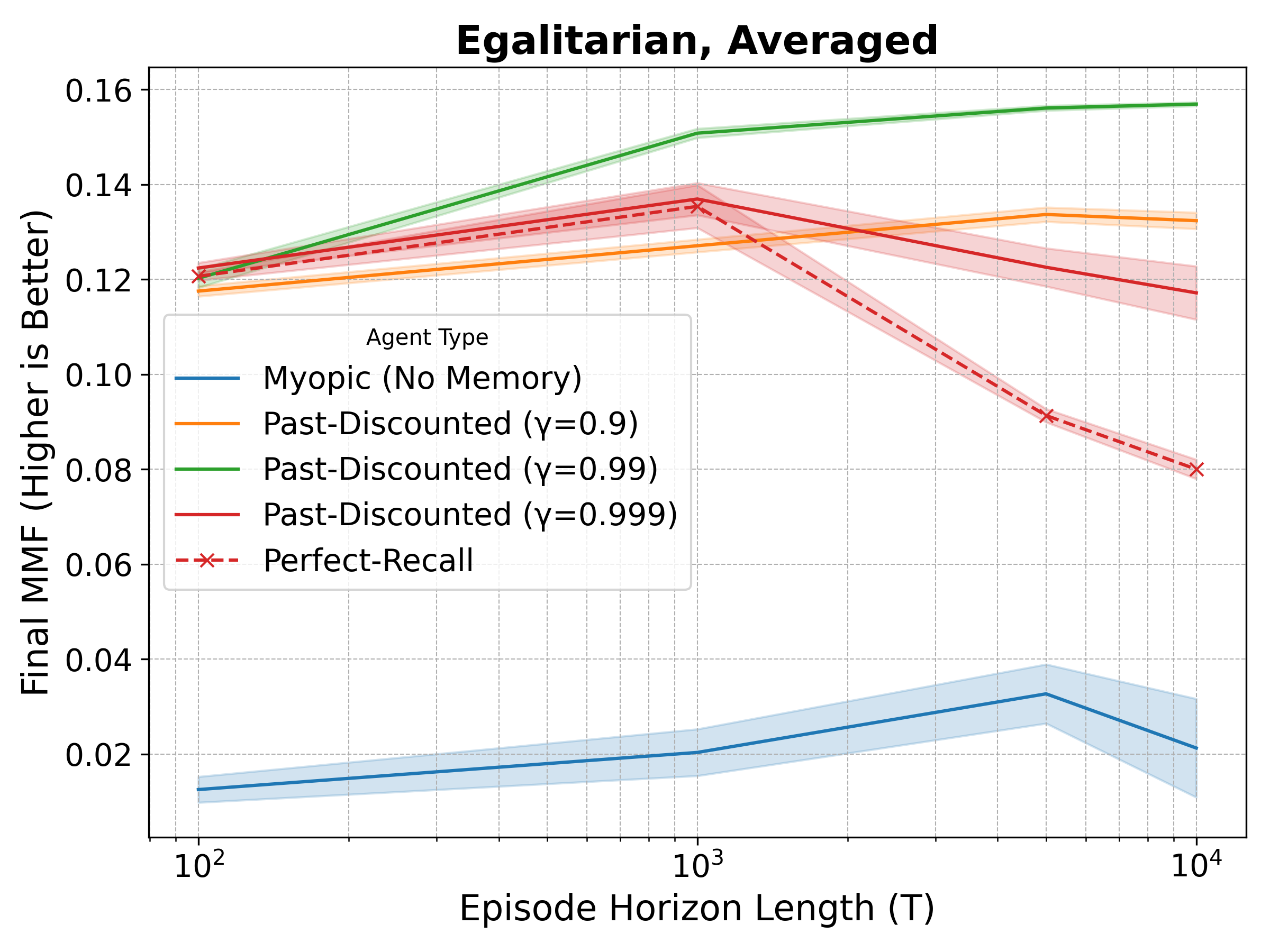}
    \includegraphics[width=0.48\linewidth]{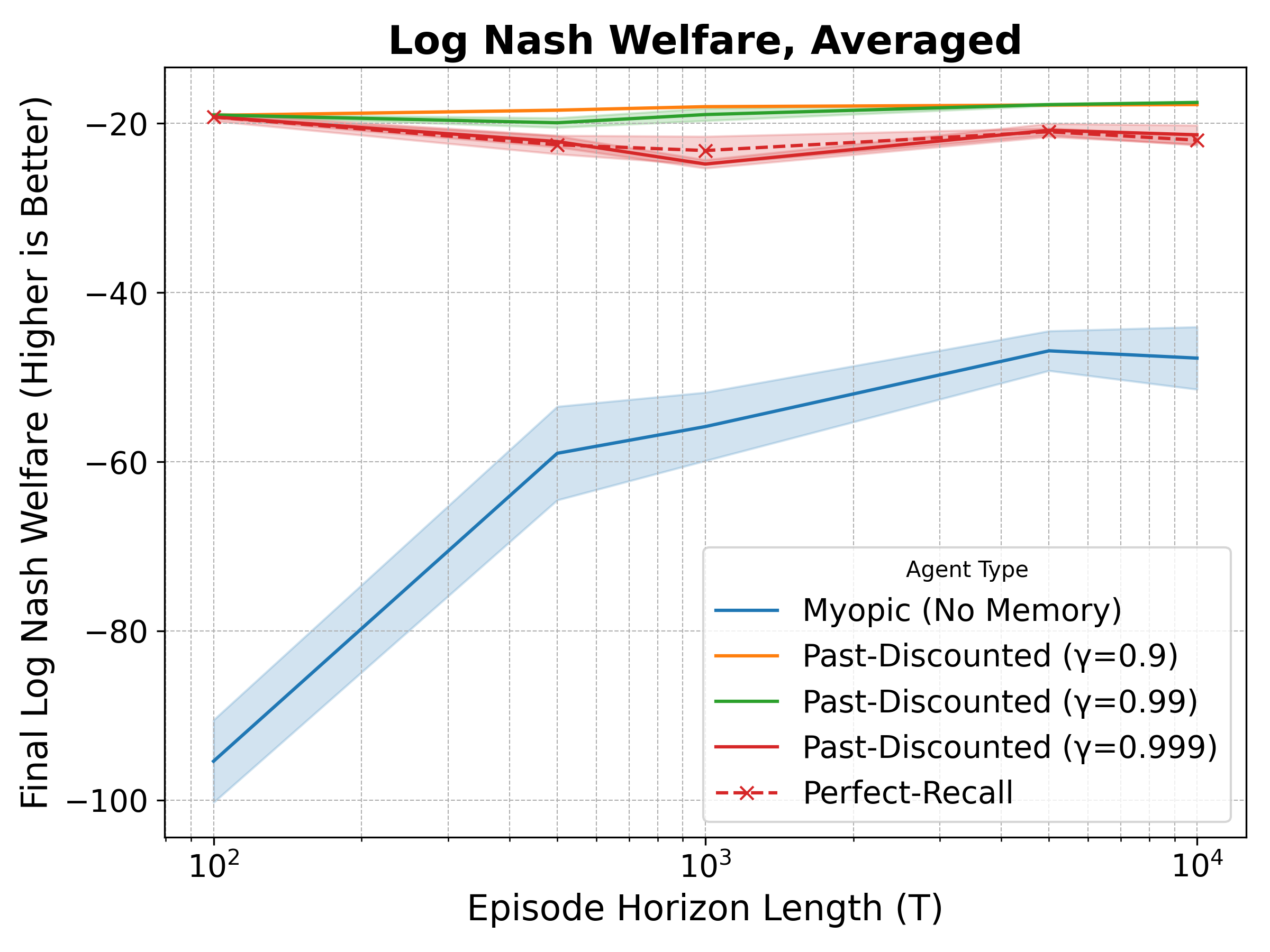}
    \caption{Final converged welfare function value for the averaged aggregation experiments as the episode length is increased. Perfect Recall's performance degrades significantly with longer horizons, while Past-Discounted methods maintain stable fairness scores.}
    \label{fig:appendix_averaged_fairness_score}
\end{figure}

\begin{figure}[h!]
    \centering
    \includegraphics[width=0.48\linewidth]{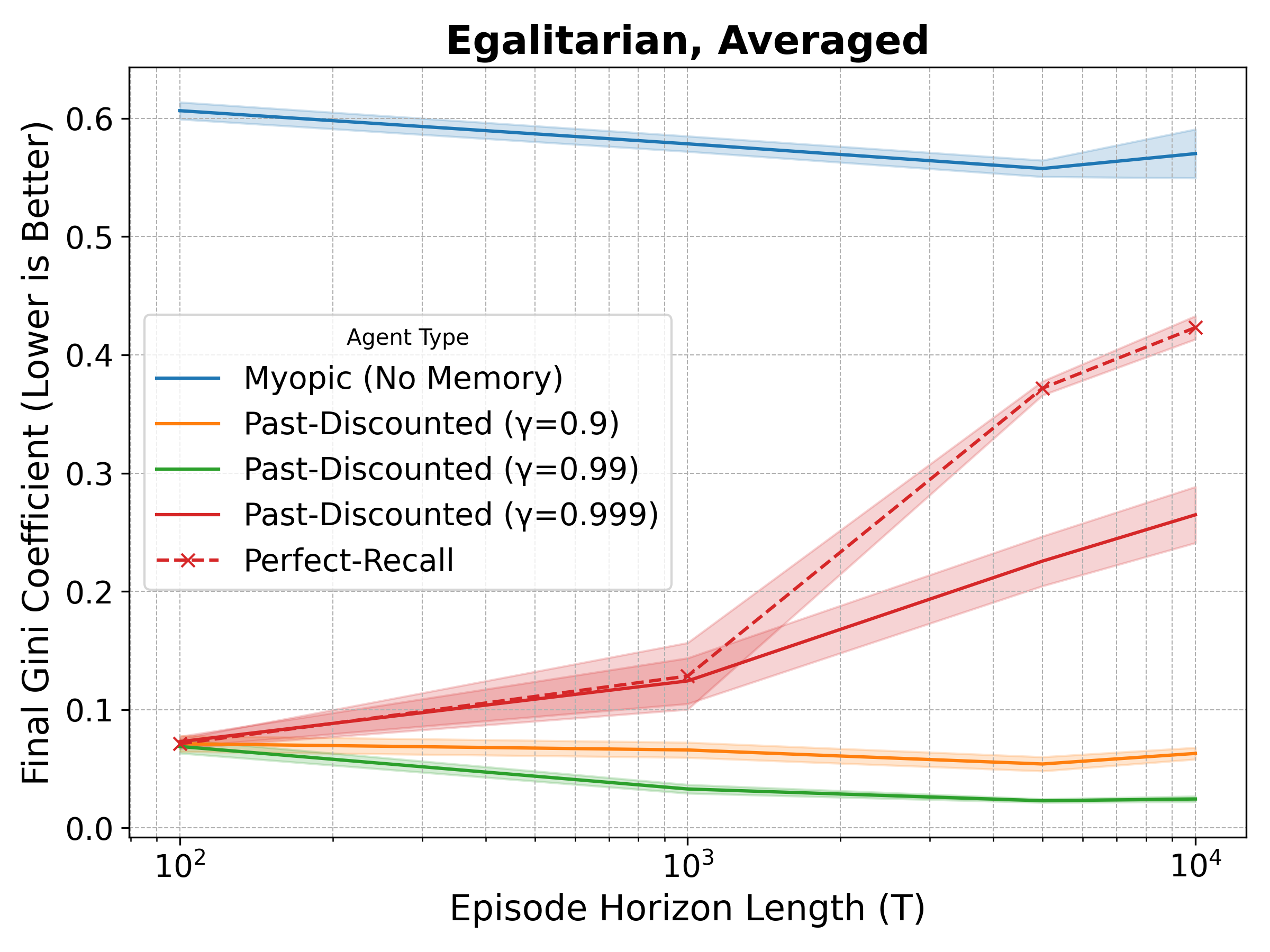}
    \includegraphics[width=0.48\linewidth]{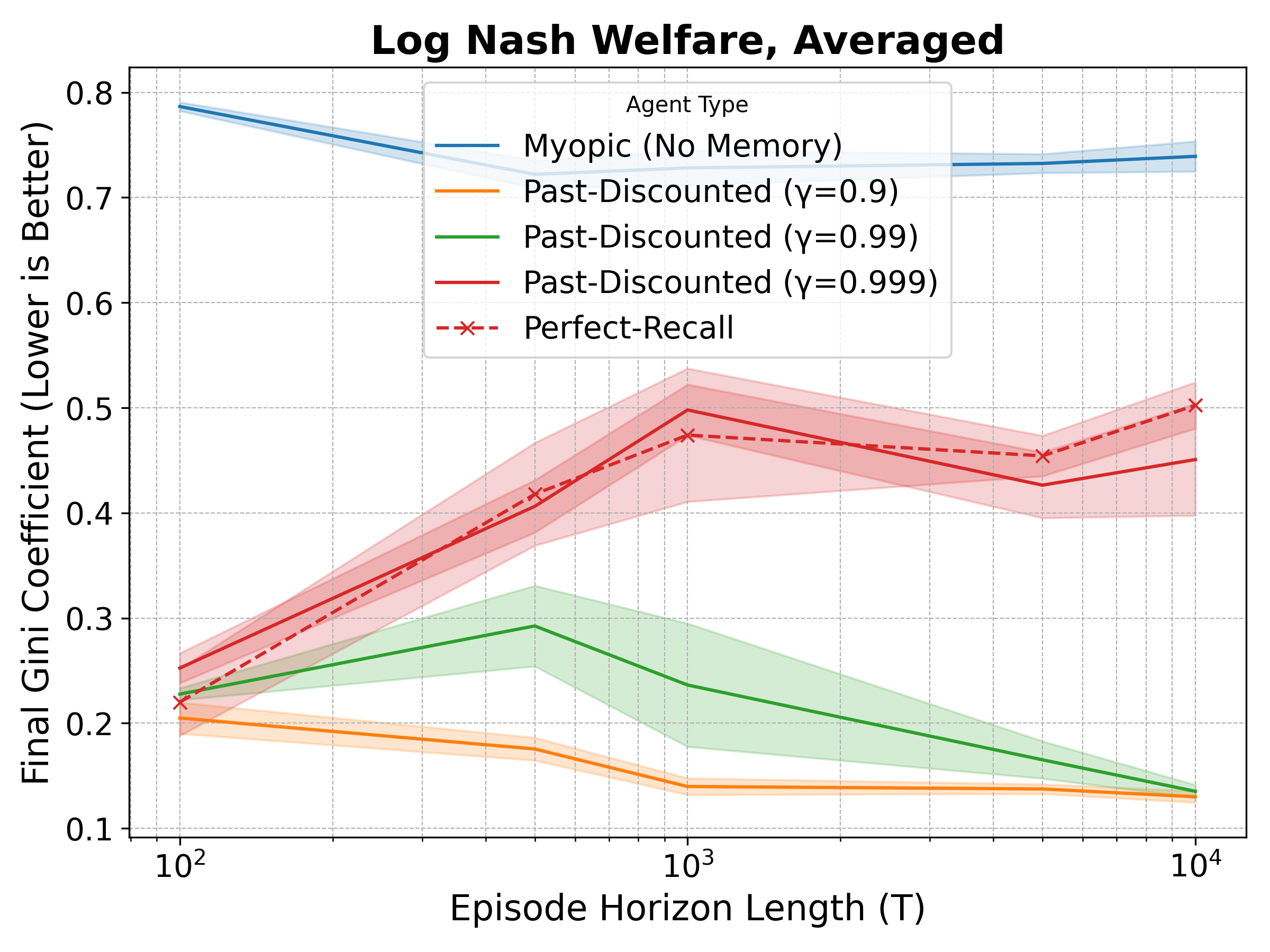}
    \caption{Gini coefficient for the \textbf{Averaged Utility} experiments as the episode length is increased. Similar to the additive case, Perfect Recall shows a marked increase in inequality at longer horizons, while Past-Discounting maintains low Gini coefficients.}
    \label{fig:appendix_averaged_gini}
\end{figure}

\section{Results for Averaged Utility Aggregation}
\label{sec:appendix_averaged_results}

To ensure our findings are robust, we conducted a parallel set of experiments using an \textbf{averaged utility aggregation} model instead of an additive one. In this configuration, the agent's fairness memory, $\mathbf{Z}^t$, represents the time-averaged utility for each agent. The core experimental setup, including the environment with advantaged agents and the PPO learning algorithm, remained identical to the additive case. The results, shown in Figures~\ref{fig:appendix_averaged_fairness_score} and~\ref{fig:appendix_averaged_gini}, confirm that the fundamental insights from our main analysis hold true for averaged utilities as well.

The \textbf{Myopic} agent, lacking any historical context, again fails to produce equitable outcomes, serving as a baseline for high inequality. The \textbf{Perfect-Recall} agent, which maintains a perfect running average of all past utilities, initially performs well on short horizons. However, as predicted by our theoretical results, this model requires an unbounded state augmentation to track the time index $t$. We observe a similar performance collapse as the episode length grows, with the agent's ability to maintain fairness degrading significantly on horizons of 5,000 steps or more. This confirms that even when utility values are bounded by averaging, the need for an ever-growing time counter makes learning intractable.

In contrast, the Past-Discounted agents using the averaged update demonstrate robust and scalable performance. By using a bounded, horizon-independent state augmentation $(\mathbf{Z}^t, d_t)$, these agents consistently learn fair policies across all tested horizons. Note that in the plot for Nash Welfare (Figure~\ref{fig:appendix_averaged_fairness_score} (bottom), it appears Perfect Recall is similar to the best methods because of the scale, as the Myopic results are extremely bad. The performance does get meaningfully worse, as can be seen in the Gini plots. The values are all negative because the sum of logs of values less than 1 (as is the case with averaged $z_i$) is always negative. 

With averaged aggregation, a new mechanism also enters the play: vanishing responsiveness. Without past-discounts, it is difficult for the Perfect Recall method to distinguish between decisions made at early time-steps versus later in the episode. Because averaging divides by the history size, the change in $\Z^t$ when $t$ is large becomes vanishingly small. Thus, with long horizons, the Perfect Recall method receives only tiny reward signals from the fairness component for a majority of the training steps. This effect is countered by using past discounting, where the effective time window makes sure transitions later on in the episode also get non-vanishing responsiveness.

These findings reinforce our main conclusion: past-discounting is a necessary mechanism for enabling tractable, long-horizon fairness, regardless of whether utilities are aggregated additively or through averaging.

\section{Detailed Training Curves}
\label{sec:appendix_training_curves}

We also provide detailed training curves for all experimental configurations discussed in the main paper. Each figure corresponds to a specific combination of welfare function (Egalitarian or Nash) and utility aggregation method (Additive or Averaged). Within each figure, we present results across various episode lengths to illustrate how each memory model performs as the time horizon increases.

\begin{figure*}[!htb]
    \centering
    \begin{subfigure}[b]{0.75\textwidth}
        \includegraphics[width=\textwidth]{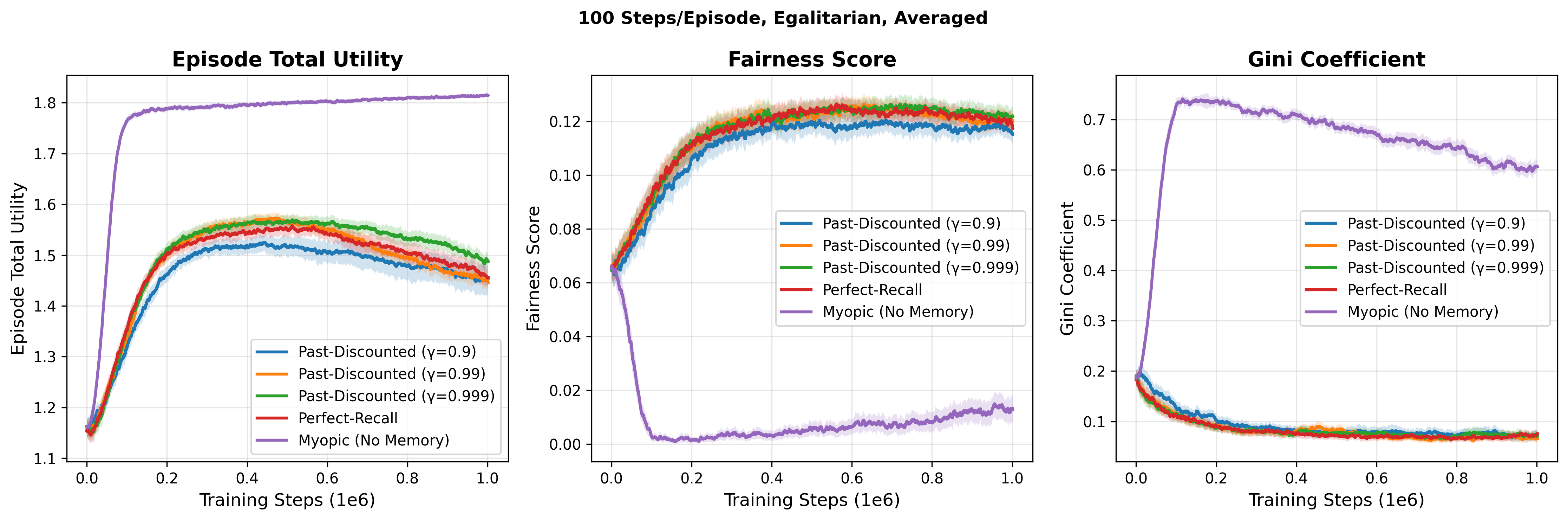}
    \end{subfigure}
    \begin{subfigure}[b]{0.75\textwidth}
        \includegraphics[width=\textwidth]{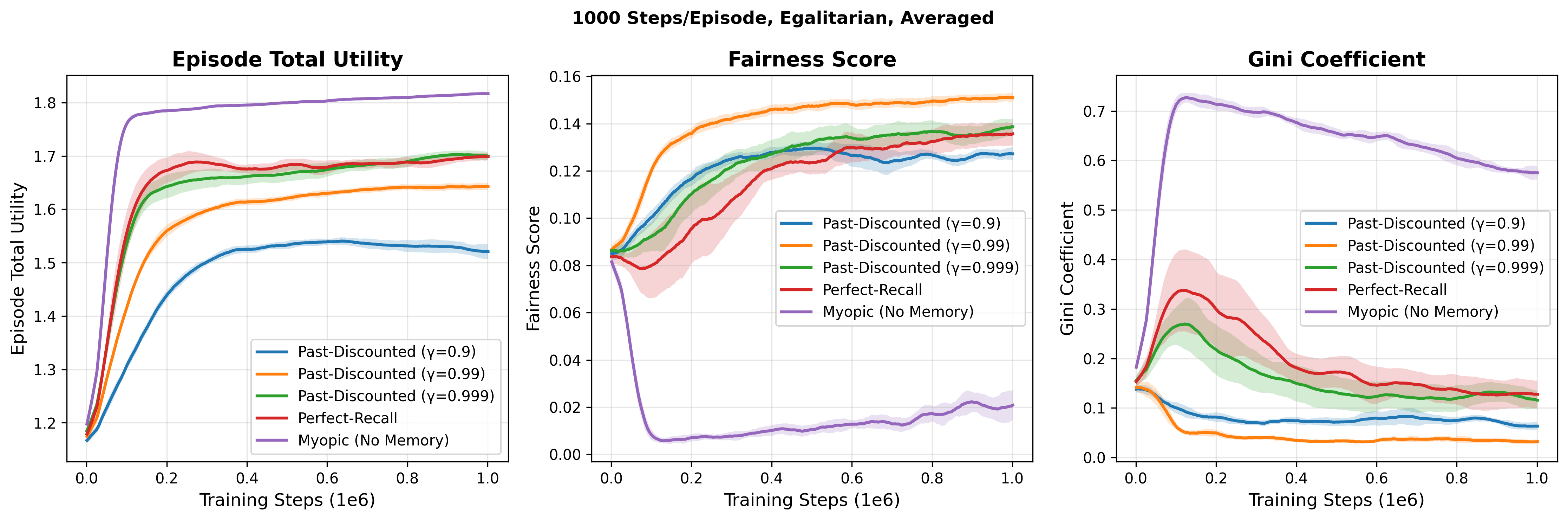}
    \end{subfigure}
    \begin{subfigure}[b]{0.75\textwidth}
        \includegraphics[width=\textwidth]{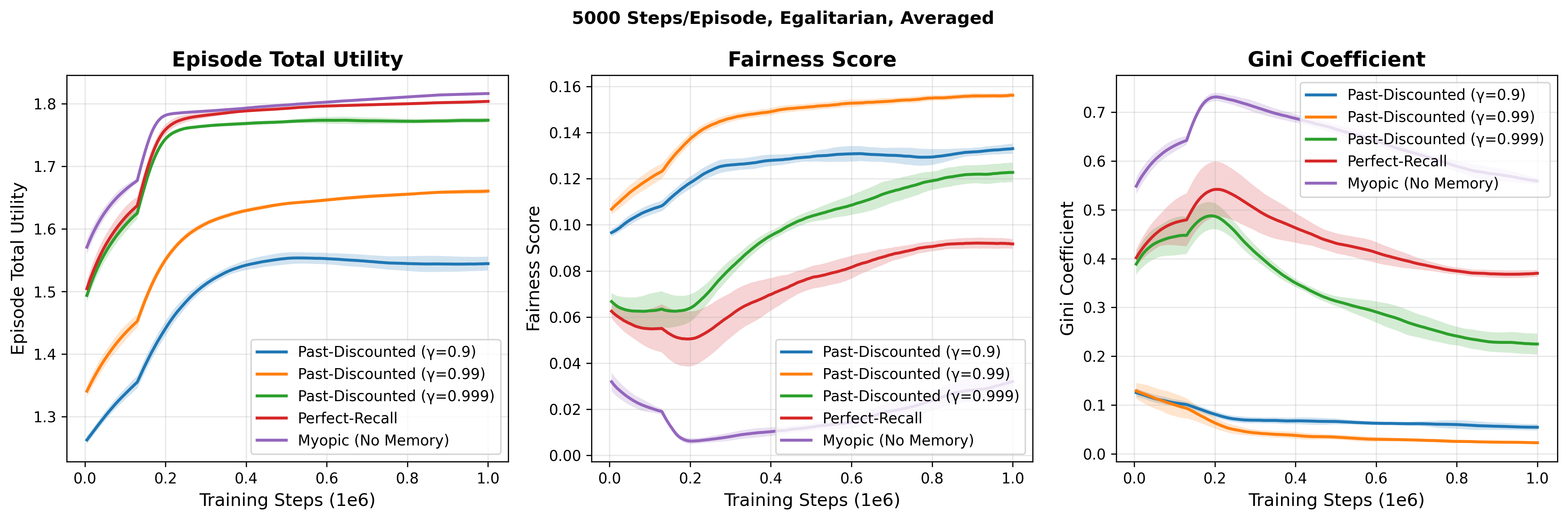}
    \end{subfigure}
    \begin{subfigure}[b]{0.75\textwidth}
        \includegraphics[width=\textwidth]{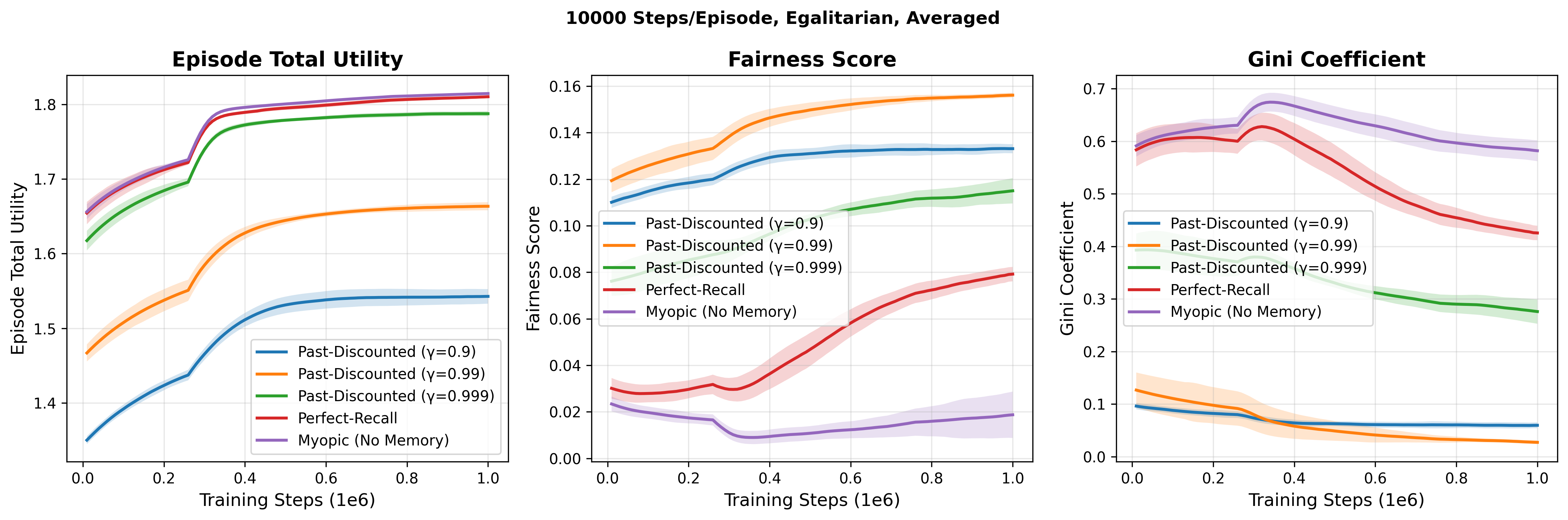}
    \end{subfigure}

    \caption{\textbf{Averaged Egalitarian Welfare:} Training curves across different episode lengths (run with $\lambda=0.999$).}
    \label{fig:appendix_averaged_egalitarian}
\end{figure*}

Each plot shows three key metrics over the course of 1 million training steps:
\begin{itemize}
    \item \textbf{Total Utility:} The sum of immediate utilities allocated at each step, averaged over the episode.
    \item \textbf{Fairness Score:} The value of the specific welfare function being optimized.
    \item \textbf{Gini Coefficient:} A measure of inequality in the final distribution of cumulative utilities, where 0 represents perfect equality.
\end{itemize}
The shaded regions in the plots represent the standard error across 10 independent runs.
For the averaged Egalitarian experiments (Figure~\ref{fig:appendix_averaged_egalitarian}), we observed that all memory models struggled to learn a stable policy with the fairness weight $\lambda=0.9$. To enable effective learning, we increased the fairness weight to $\lambda=0.999$ for these runs, which provided a stronger, more consistent signal for the agent. All other experiments were conducted with $\lambda=0.9$.

These are the general trends to observe in all these experiments:
\begin{itemize}
    \item As the horizon is increased, the perfect recall method starts exhibiting worse performance in terms of fairness.
    \item The myopic policy converges to a utility of 1.8 per time step, indicating that it consistently picks the advantaged agents whose average need would be 0.9.
    \item The effective window's effect is visible based on how closely past-discounted experiments track perfect recall. We expect each method to be close to perfect recall for horizons of similar or smaller scale to the effective time window of the selected $\gamma_p$.
    \item In some cases, it looks like the poorly performing methods are catching up as the steps approach $1e-6$. This further adds to the strengths of the past-discounting approach: the increased complexity of the state-space is making the problem much harder to learn from, making sample complexity worse. So even if Perfect Recall should be able to learn given sufficient time, this is very long compared to the steps needed for past-discounted agents, and gets longer as the horizon increases. 
\end{itemize}


\begin{figure*}[!htb]
    \centering
    \begin{subfigure}[b]{0.75\textwidth}
        \includegraphics[width=\textwidth]{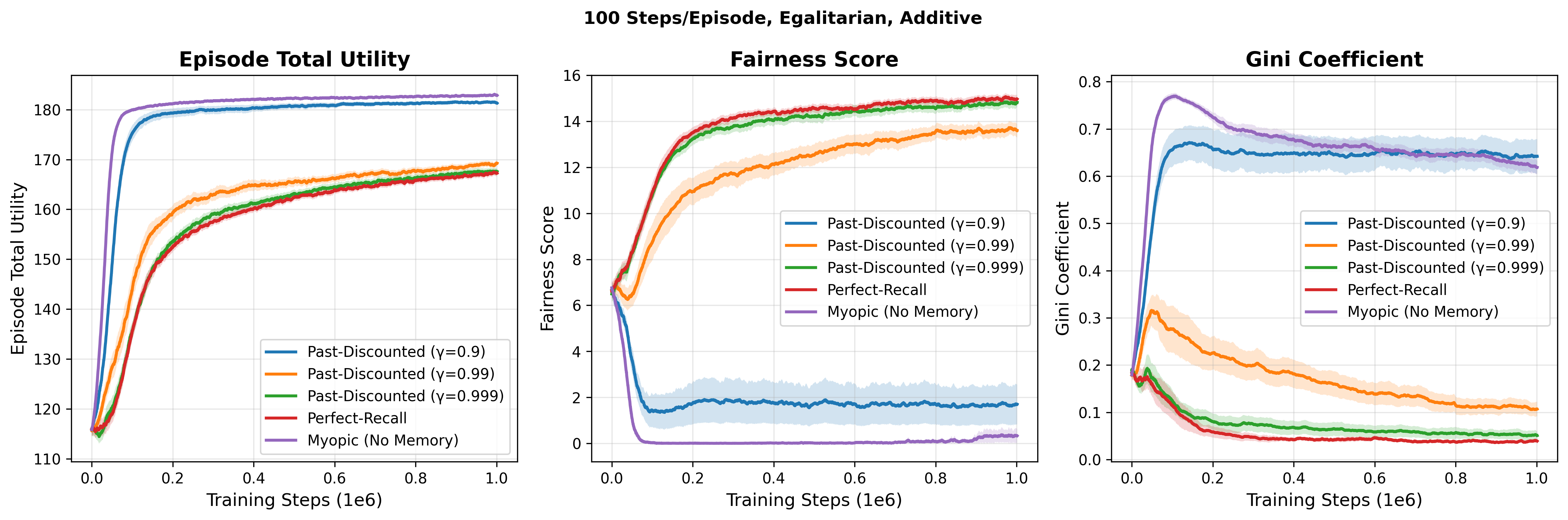}
    \end{subfigure}
    \begin{subfigure}[b]{0.75\textwidth}
        \includegraphics[width=\textwidth]{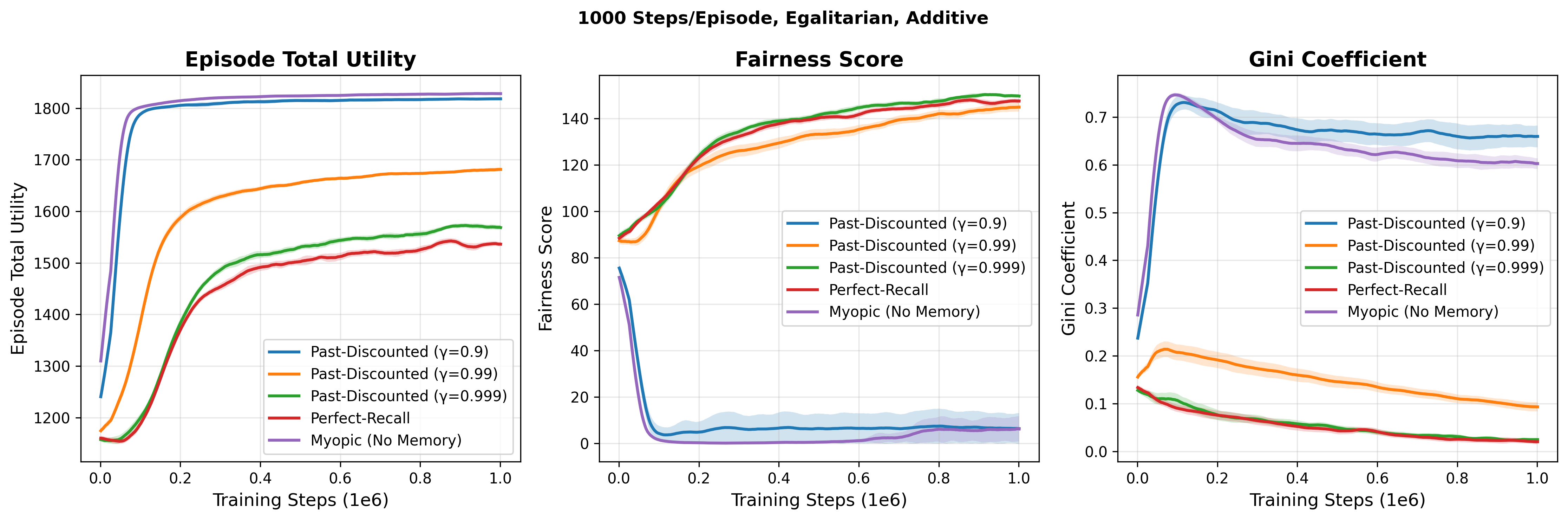}
    \end{subfigure}
    \begin{subfigure}[b]{0.75\textwidth}
        \includegraphics[width=\textwidth]{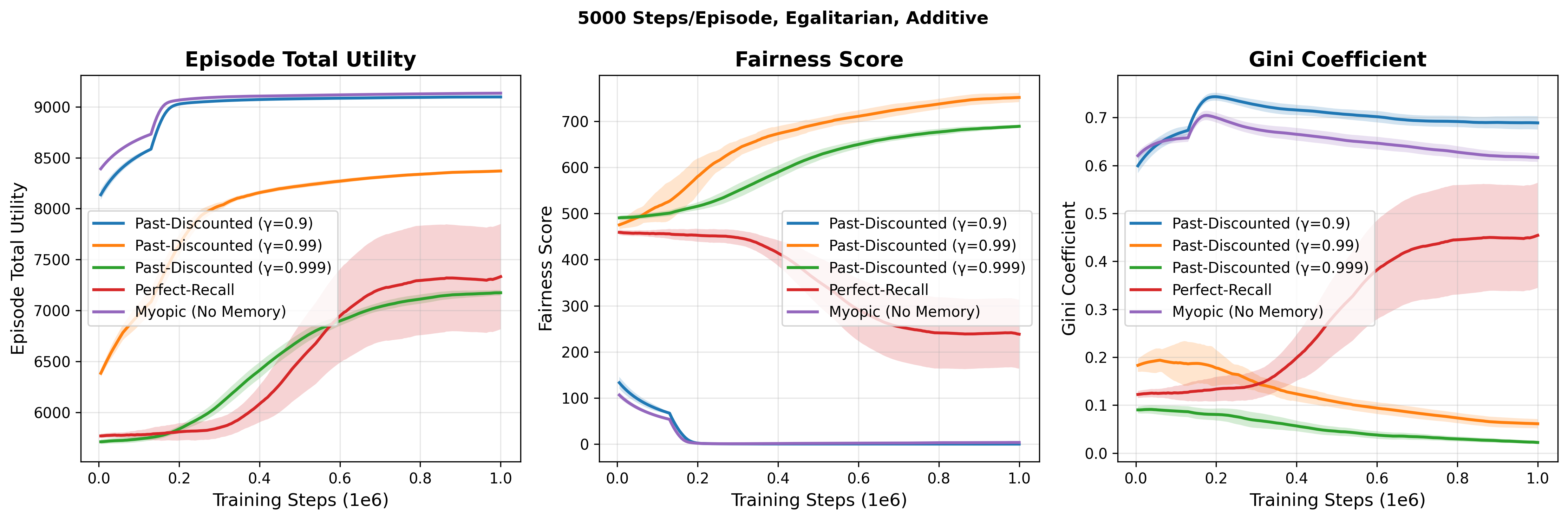}
    \end{subfigure}
    \begin{subfigure}[b]{0.75\textwidth}
        \includegraphics[width=\textwidth]{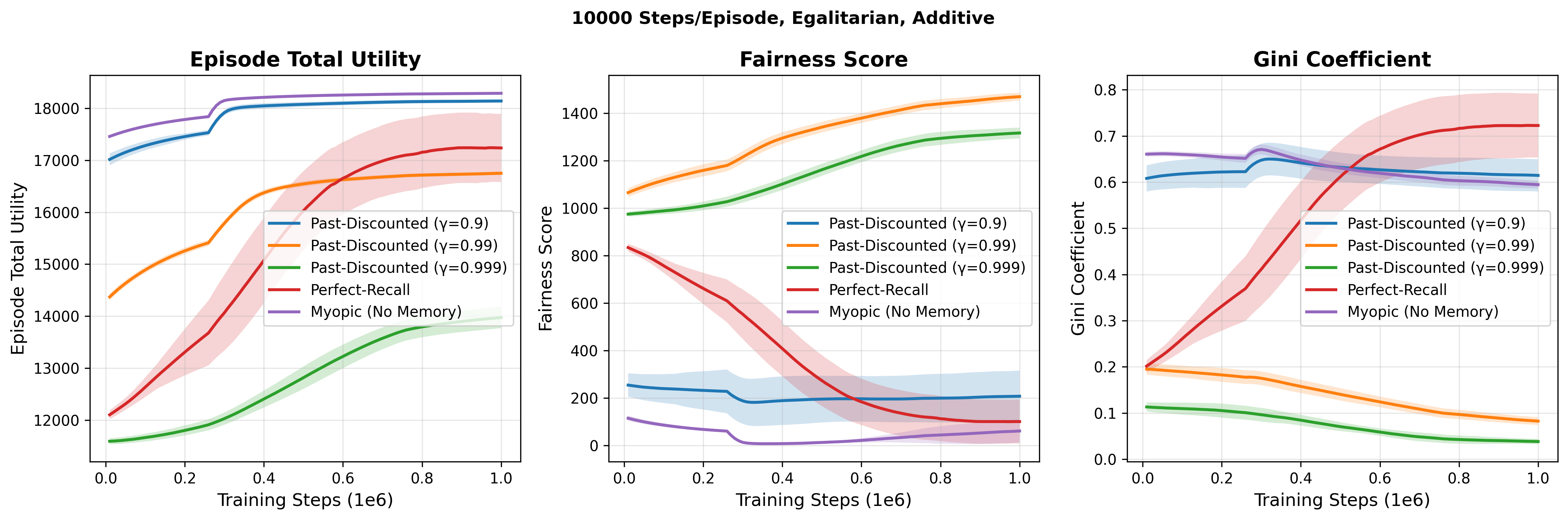}
    \end{subfigure}
    
    \caption{\textbf{Additive Egalitarian Welfare:} Training curves across different episode lengths.}
    \label{fig:appendix_additive_egalitarian}
\end{figure*}

\clearpage

\begin{figure*}[!htb]
    \centering
    \begin{subfigure}[b]{0.75\textwidth}
        \includegraphics[width=\textwidth]{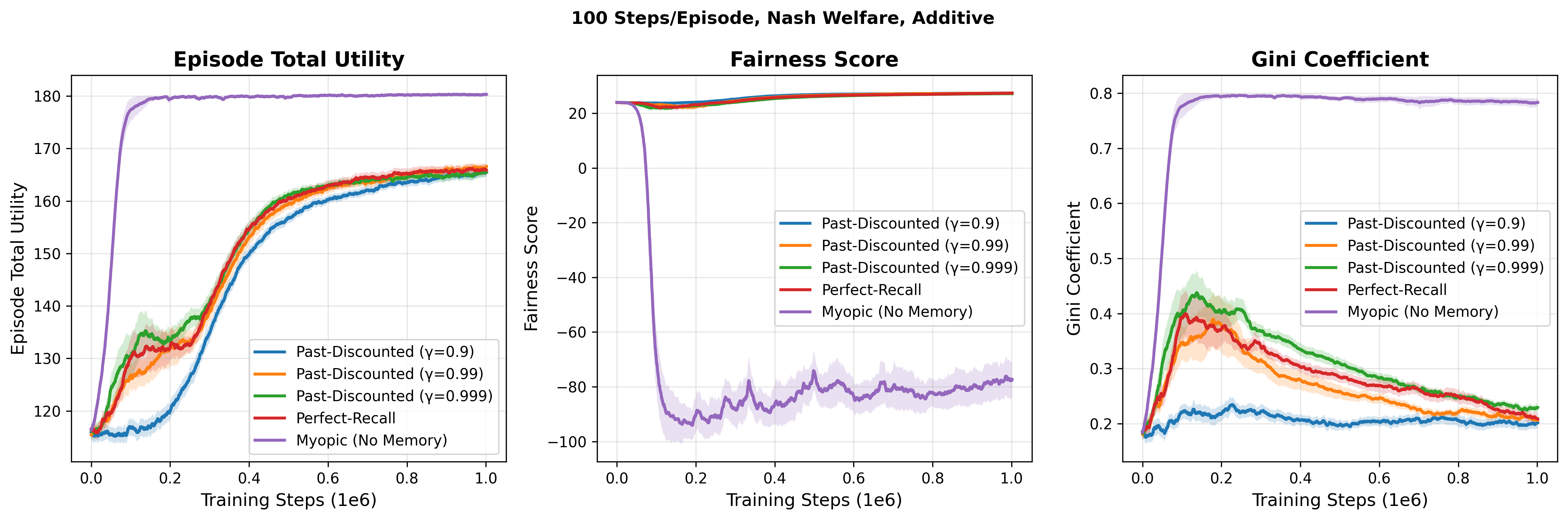}
    \end{subfigure}
    \begin{subfigure}[b]{0.75\textwidth}
        \includegraphics[width=\textwidth]{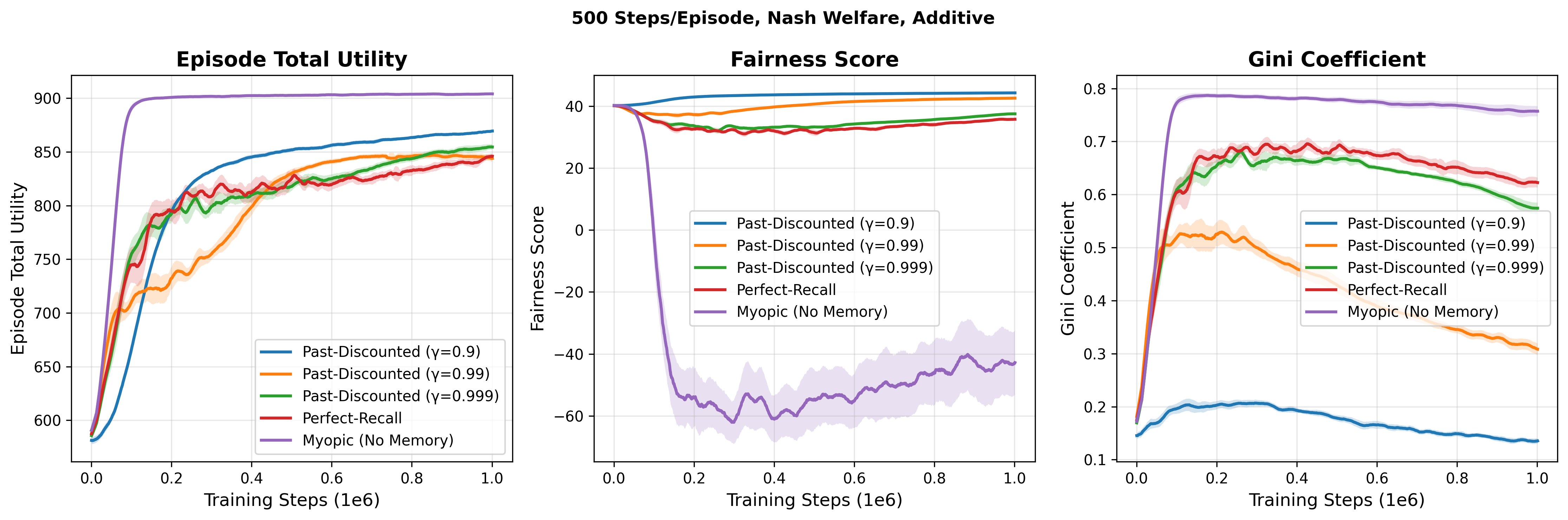}
    \end{subfigure}
    \begin{subfigure}[b]{0.75\textwidth}
        \includegraphics[width=\textwidth]{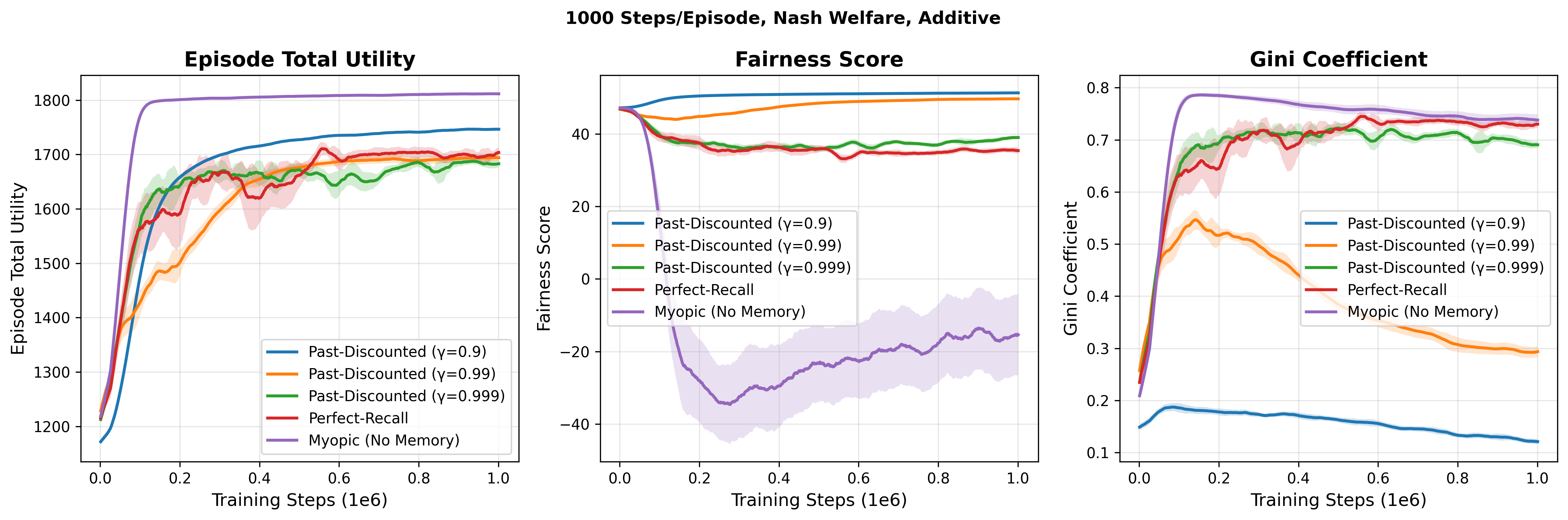}
    \end{subfigure}
    \begin{subfigure}[b]{0.75\textwidth}
        \includegraphics[width=\textwidth]{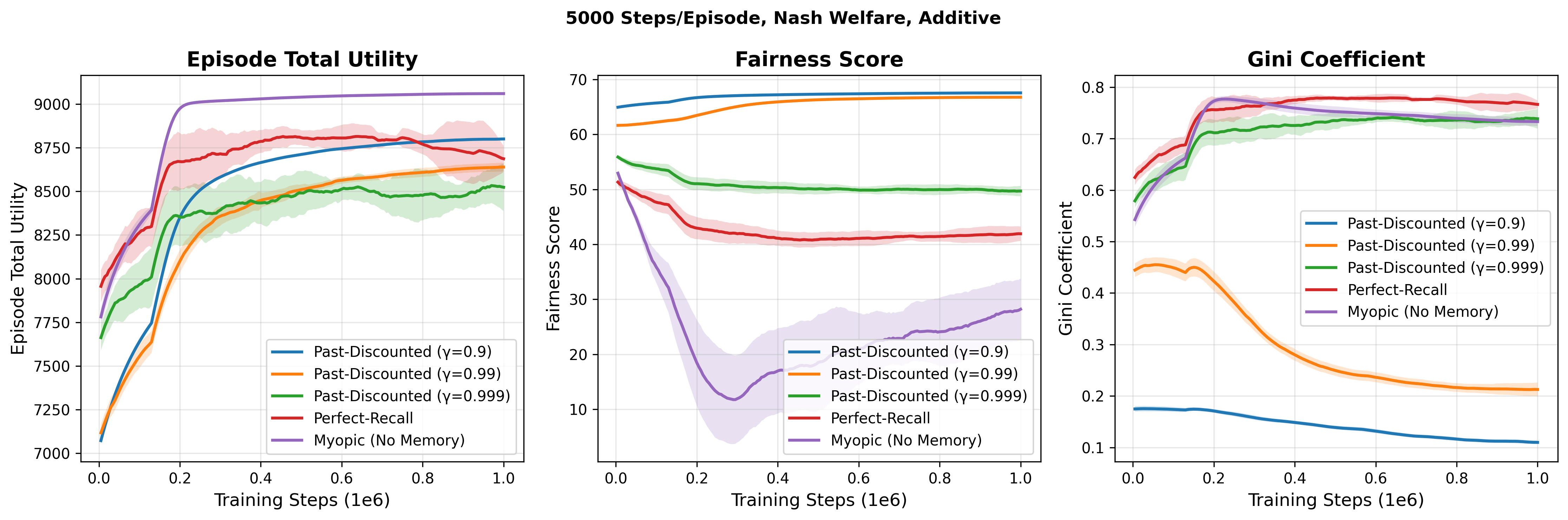}
    \end{subfigure}
    \begin{subfigure}[b]{0.75\textwidth}
        \includegraphics[width=\textwidth]{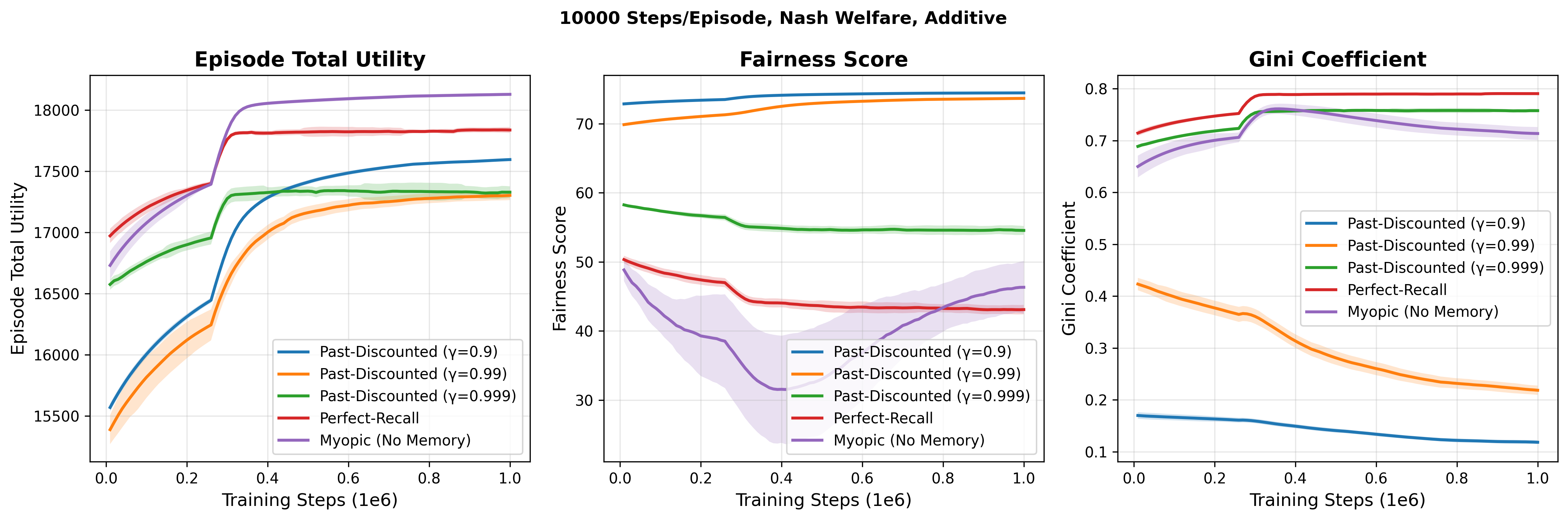}
    \end{subfigure}
    
    \caption{\textbf{Additive Nash Welfare:} Training curves across different episode lengths.}
    \label{fig:appendix_additive_nash}
\end{figure*}

\clearpage

\clearpage
\begin{figure*}[!htb]
    \centering
    \begin{subfigure}[b]{0.75\textwidth}
        \includegraphics[width=\textwidth]{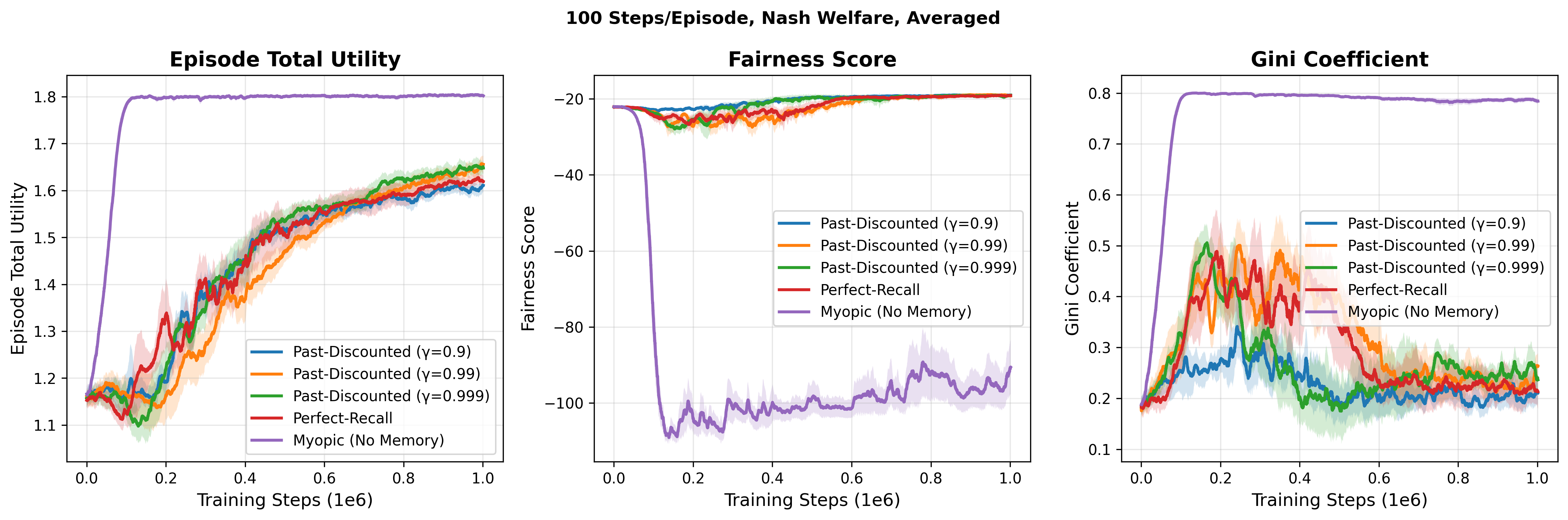}
    \end{subfigure}
    \begin{subfigure}[b]{0.75\textwidth}
        \includegraphics[width=\textwidth]{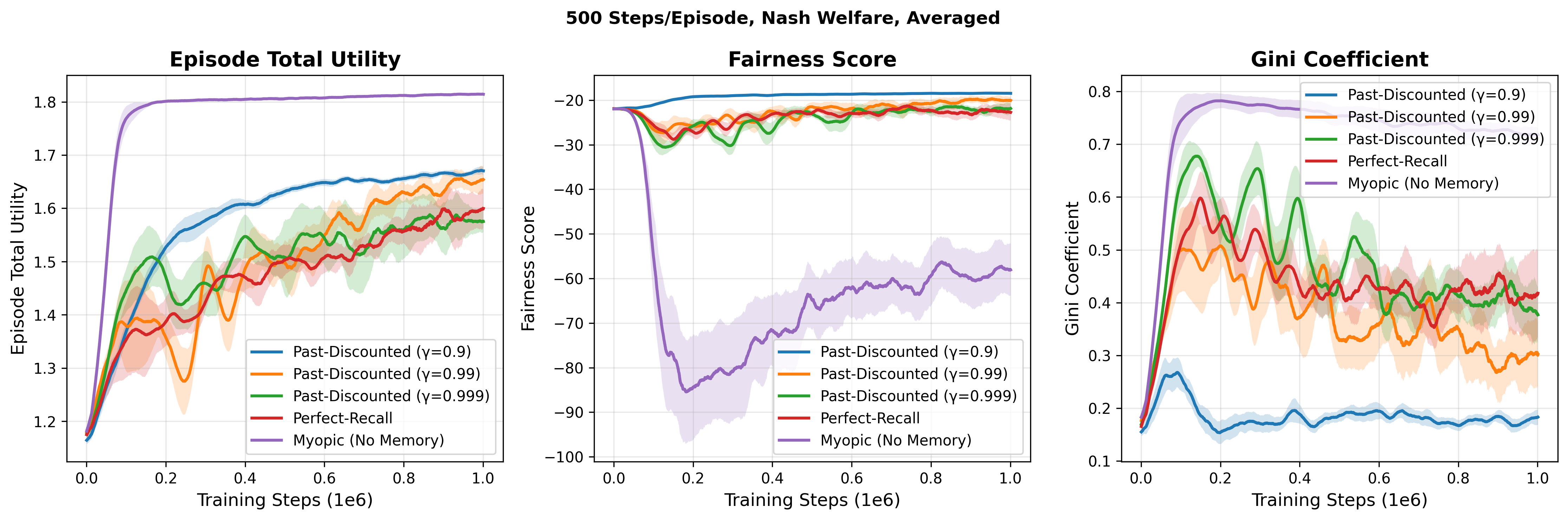}
    \end{subfigure}
    \begin{subfigure}[b]{0.75\textwidth}
        \includegraphics[width=\textwidth]{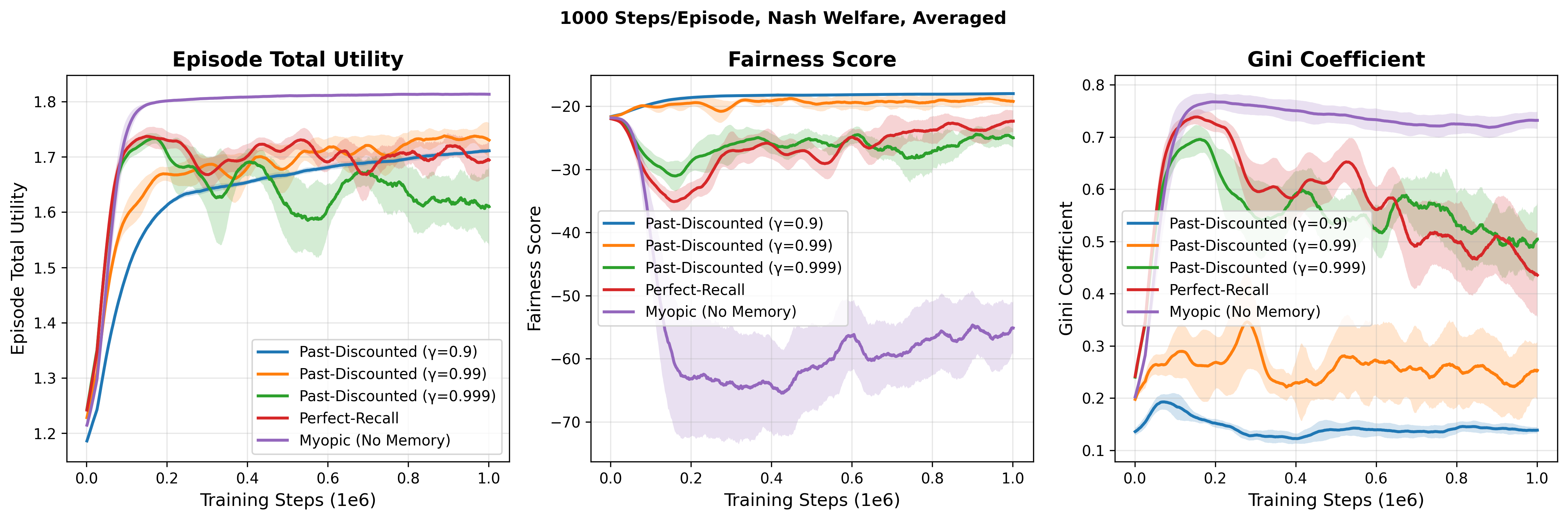}
    \end{subfigure}
    \begin{subfigure}[b]{0.75\textwidth}
        \includegraphics[width=\textwidth]{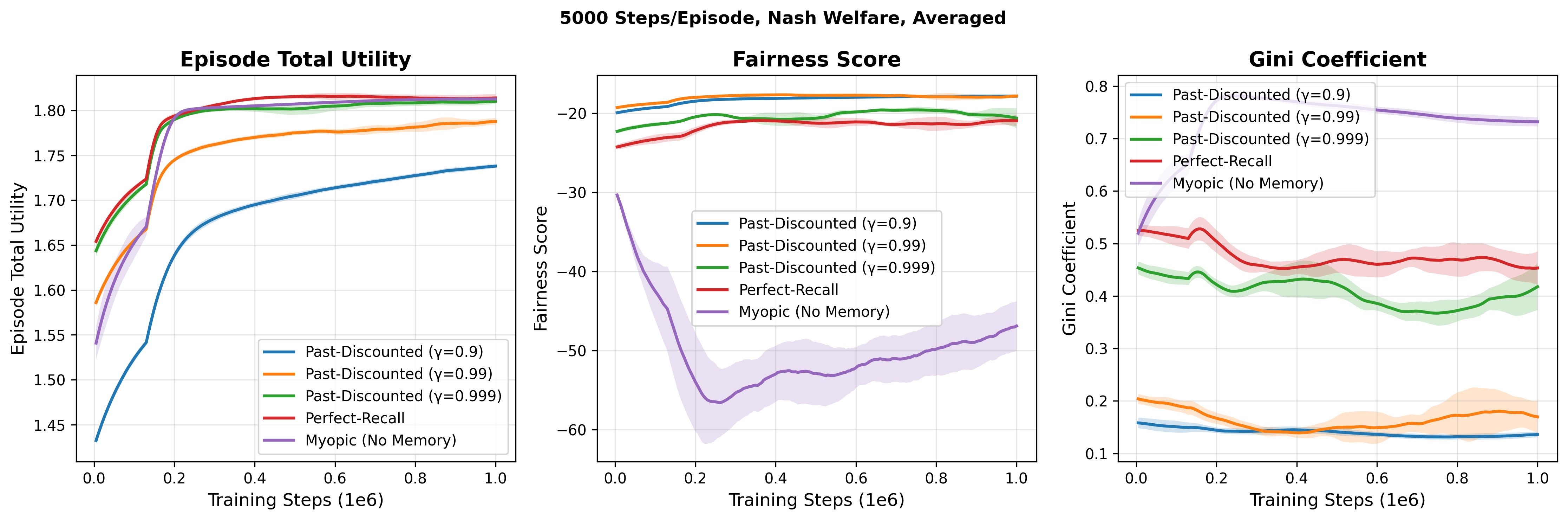}
    \end{subfigure}
    \begin{subfigure}[b]{0.75\textwidth}
        \includegraphics[width=\textwidth]{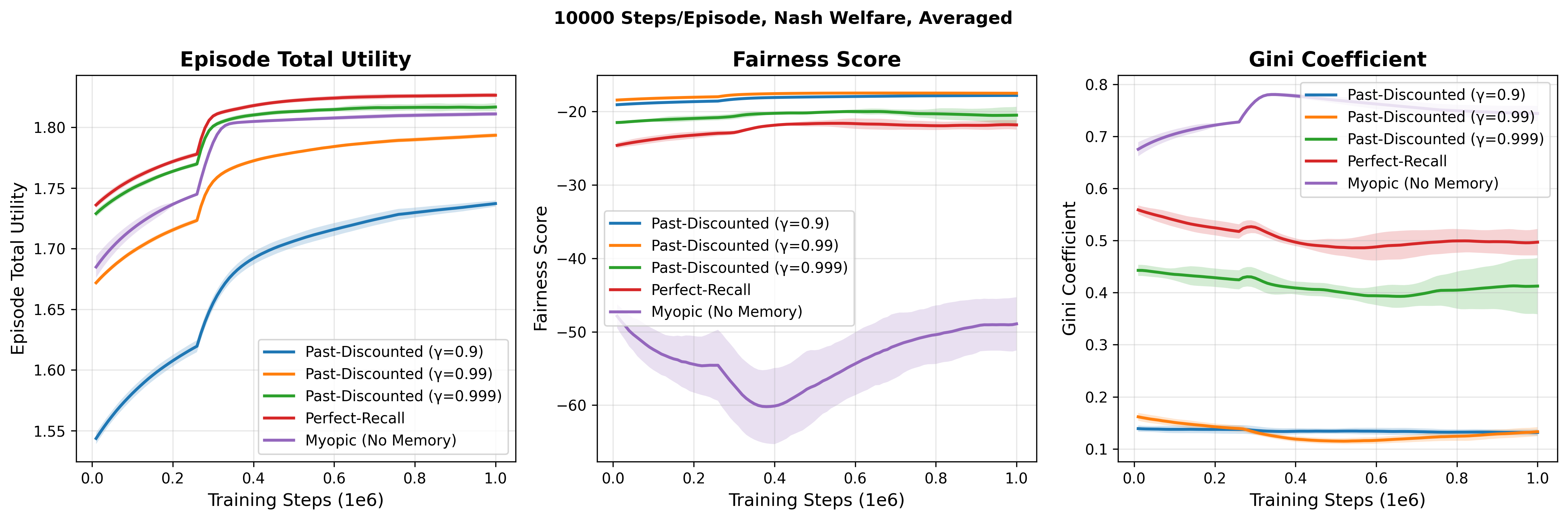}
    \end{subfigure}
    
    \caption{\textbf{Averaged Nash Welfare:} Training curves across different episode lengths.}
    \label{fig:appendix_averaged_nash}
\end{figure*}